# Smart Mobility Ontology: Current Trends and Future Directions

**Ali Yazdizadeh · Bilal Farooq**

**Abstract** Ontology is the explicit and formal representation of the concepts in a domain and relations among them. Transportation science is a wide domain dealing with mobility over various complex and interconnected transportation systems, such as land, aviation, and maritime transport, and can take considerable advantage from ontology development. While several studies can be found in the recent literature, there exists a large potential to improve and develop a comprehensive smart mobility ontology. The current chapter aims to present different aspects of ontology development in general, such as ontology development methods, languages, tools, and software. Subsequently, it presents the currently available mobility-related ontologies developed across different domains, such as transportation, smart cities, goods mobility, sensors. Current gaps in the available ontologies are identified, and future directions regarding ontology development are proposed that can incorporate the forthcoming autonomous and connected vehicles, mobility as a service (MaaS), and other disruptive transportation technologies and services.

Ali Yazdizadeh
LiTrans, Ryerson University
E-mail: ali.yazdizadeh@ryerson.ca

Bilal Farooq
LiTrans, Ryerson University
E-mail: bilal.farooq@ryerson.ca

## 1 Introduction

Ontology is the explicit formal representation of the concepts in a domain, the relations between them, and their properties and constraints (Gruber et al., 1993). Ontology, as a discipline of philosophy, explains the nature of existence and has its roots in Aristotle and Plato studies on “metaphysics” (Welty and Guarino, 2001). However, the word *ontology* originated from two Greek words: ontos (being) and logos (word), and conceived for the first time during the Sixteen century by German philosophers (Welty and Guarino, 2001). From then till the mid-twentieth, ontology evolved mainly as a branch of philosophy. However, with the advent of Artificial Intelligence since the 1950s, researchers perceived the necessity of ontology to describe a new world of intelligent systems (Welty and Guarino, 2001). Moreover, with the development of the World Wide Web in the 1990s, ontology development got to be common among different domain specialists to define and share the concepts and entities in their fields on the Internet (Noy et al., 2001). During the last three decades, ontology development studies have evolved and shifted from theoretical issues of ontology to practical implications of the use of ontology in real-world, large-scale applications (Noy et al., 2001).

Nowadays, ontology development focuses mainly on defining machine interpretable concepts and their relationships in a domain. However, ontology development also pursues other goals, such as providing a common conceptualization of the domain on which different experts agree, (Métral and Cutting-Decelle, 2011) and enable them to reuse the domain knowledge (Noy et al., 2001). It also enables researchers to easily analyze the domain knowledge and eloquently express the domain assumptions. Furthermore, ontology facilitates sharing a common understanding of the structure of concepts among people or software agents (Noy et al., 2001).

Transportation science is a wide domain dealing with a variety of complex and interconnected transportation systems, such as land, aviation, and maritime transport. Figure 1 shows the different disciplines and fields of science involved in smart transportation. The image has been generated by searching for publications and conference papers that include the “Smart Mobility” phrase on Web of Science. The image shows around 20 different fields of science involved in smart mobility. Such a multidisciplinary domain can take a considerable advantage from ontology development. Several reasons justify the demand for a comprehensive mobility ontology. First, since the last two decades, the technological innovations in data collection techniques, from smartphone and WiFi technologies to the different roadside, on-vehicle and on-body sensors, monitoring cameras, etc., have provided transportation/mobility scientists and planners with huge amounts of data, summarized under the term “Big Transportation Data”, which consequently have boosted the demand for cloud platforms and storage systems during last years (Badu-Marfo et al., 2019). Such cloud platforms and storage systems usually suffer from heterogeneity issues, as they are required to encompass a variety of services and data, each of which possesses different features, definitions, and details without rel-evant standards and conventions to enable interoperability between various environments (Al-Sayed et al., 2019). Second, in the realm of mobility, the data is collected by different public and private parties and often stored and accessed in separate datasets that are extremely hard to organize and form into an integrated management or monitoring system.

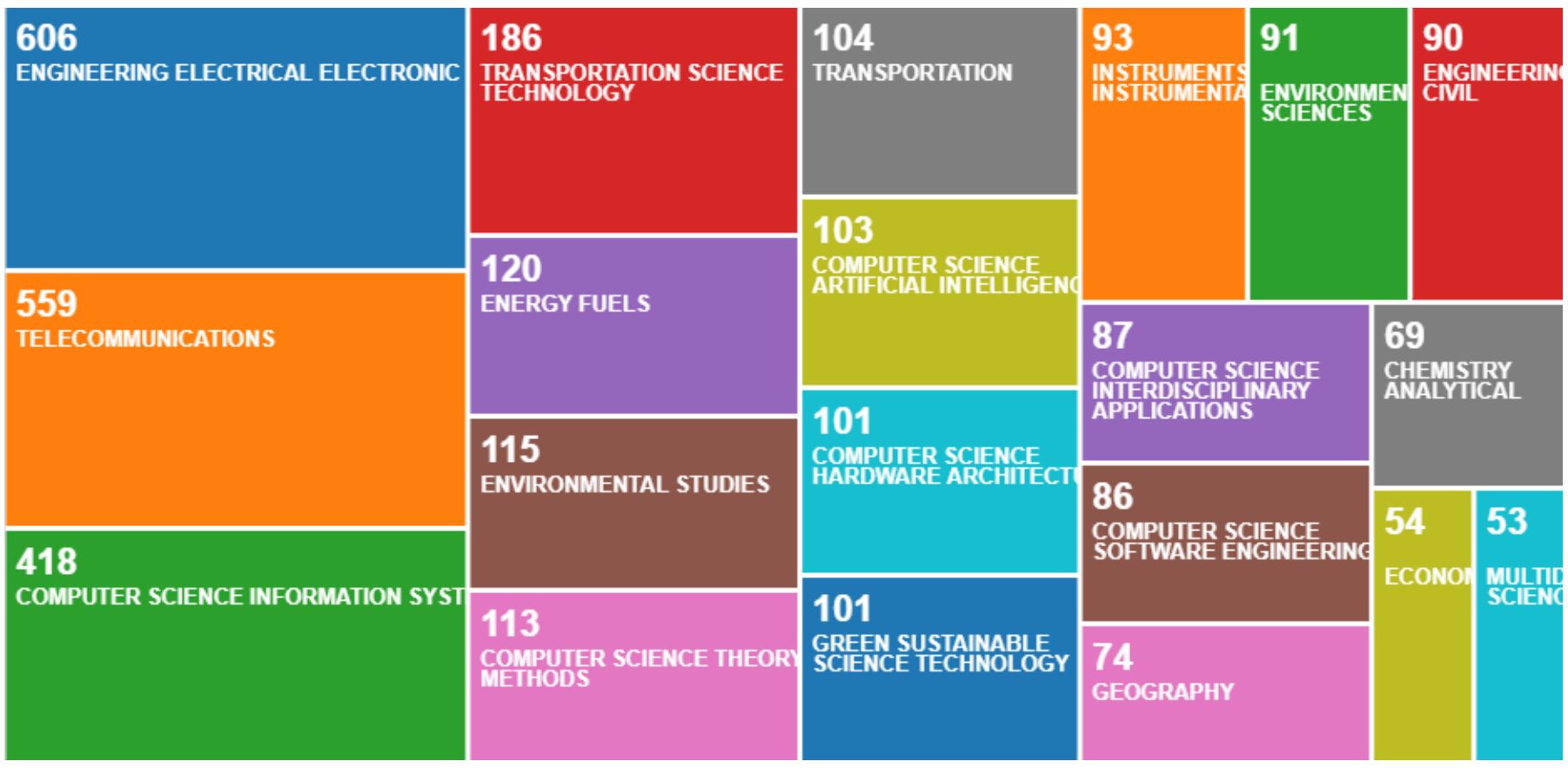

**Fig. 1** Different fields of science and disciplines involved in transportation studies. (generated by Web Of Science search engine)

As ontology has first been employed in the artificial intelligence domain to conceptualize some real-world elements (Falquet et al., 2011), it is a rather new concept in the transportation/mobility domain. While several studies exist in the literature, there exists a large potential for improving and developing mobility ontology. The major goals of this study are: to present the different aspects of ontology development, review the existing mobility ontologies, and identify the gaps in the current literature to address future development. The chapter begins by explaining different components of ontology in Section 2. Afterwards, Section 3 describes different ontology classification approaches. Next, Section 4 explains various ontology development approaches and points to some examples in mobility ontology. Section 5 gives a brief background on different languages for ontology development. Next, Section 6 explains the general procedure and steps for the development of an ontology.

Afterward, Section 7 categorizes different sectors in the transportation domain to identify the parts covered in the current literature and the parts requiring more attention. It also reviews the current ontologies in the domain. Afterward, Section 8 explains the gaps in the literature, specifically in disruptive transportation technologies and services, and discusses possible solutions to fill these gaps in mobility ontology development. Finally, Section 9 concludes and summarizes the future directions.

## 2 Ontology Components

Here we elaborate on the main components in an ontology, regardless of its formality and domain granularity. Every ontology is at least composed of three main components (Noy et al., 2001):

- **Concepts (classes)**
- **Properties** (also called **roles** or **slots**) of each concept
- **Restrictions** on properties (also called restrictions on slots or **facets**)

*Classes* represent the entities or concepts in a domain and are the core elements of any ontology. For example, in the domain of transportation, a class of *vehicle* defines all types of vehicles. Specific *vehicles* are instances of this class. For example, *bus* can be considered as an instance of the class *vehicle*. In an ontology a class can have *sub-classes*. For example, we can divide vehicles into personal and public vehicles. Alternatively, we can divide vehicles into motorized and unmotorized vehicles.

Each class in ontology has some attributes or characteristics. For example, an instance of class *vehicle* has a Name, Brand, and Fuel Type that describe it. In ontology, the characteristics or attributes of a vehicle are referred to as *Properties*. Indeed, the classes are described by their *Properties* in more detail. *Restrictions* on properties or *facets* define the data type of properties, the range and domain of values a property can take, also the number of values (referred to as cardinality of properties) a property can have. For example, in vehicle class, the *Fuel Type* property is of data type "String". Moreover, *Fuel Type* can only take specific types of fuel, for example, Gasoline, Diesel, Electricity, Ethanol, Biodiesel, and Hydrogen. Furthermore, regarding the cardinality, a vehicle can run on just one type of fuel, such as gasoline, or on two types, like gasoline and electricity. Hence, the minimum cardinality of *Fuel Type* in this case, can be defined as 1, and the maximum of it equals 2. However, some properties may only accept one value, for example, the *color* property of the class *vehicle*.

From a practical point of view, ontology development mainly focuses on the following four steps (Noy et al., 2001):

- Define the classes
- Define the taxonomic hierarchy of classes
- Define properties and their allowed values
- Specify the instances and set the values for the properties of each instance

The term "taxonomy hierarchy" refers to the sub-class/super-class hierarchy in an ontology and will be explained in Section 4. Also, Section 6 explains the above steps in detail and describes a framework for ontology development. Before going to more details about ontology development approaches and procedures, the next section explains different types of ontology and how they are classified in the literature.

## 3 Ontology Classification

Ontologies can be classified based on different dimensions (Roussey et al., 2011):

- Language expressivity and formality
- Scope of the ontology, or domain granularity

Based on language expressivity and formality, there are four types of ontologies, ranked from less formal languages to more formal ones (Roussey et al., 2011):

- **Information ontologies**: used only by humans, the goal of information ontologies is to clarify and organize the ideas and plans in the development of a project using visual languages, i.e., diagrams and sketches, such as a mind map.
- **Linguistic/Terminological ontologies**: focus on terms and their relationships, linguistic ontologies can be any type of glossaries, dictionaries, lexical databases, Web metadata, etc. For example, the Resource Description Framework (RDF) is a general-purpose language to represent concepts, terminology, and information on the Internet (Roussey et al., 2011).
- **Software ontologies**: used for software development projects, provide the schema for databases, and data manipulation to guarantee data consistency. During the software design procedure, usually, a conceptual modeling language, such as Unified Modeling Language (UML), is used.
- **Formal ontologies**: usually developed using formal logic (ex. first-order logic or description logic) to describe the rules about how to define the concepts and relationships. The most well-known formal language is OWL, which will be introduced in Section 5.

With respect to scope, or domain granularity, ontologies fall into four different categories (Roussey et al., 2011):

- **Application/Local ontologies**: are developed according to a user or a developer's viewpoint to give a representation of the particular model of a domain. Application/Local ontologies can be deemed as a task ontology, combined with domain ontology, to define concepts and their relationships regarding a specific purpose of an application.
- **Core ontologies**: can be considered as a basic and minimal ontology that defines the minimal concepts in a domain that are essential to understand the other concepts.
- **Domain ontologies**: represent concepts and relationships regarding a specific domain of the real-world, for example, transit network. Domain ontologies show how a group of users/experts perceive a specific domain, what are the main concepts in the domain, and what their properties are.
- **General ontologies**: define concepts and relationships related to a huge area of knowledge.
- **Foundation ontologies**: are top-level generic ontologies applicable to various domains (Roussey et al., 2011). Every core or domain ontology usually includes a foundation ontology, such as Basic Formal Ontology (BFO) (Arp et al., 2015). Domain ontologies developed based on the same foundation ontology are more interoperable and can be easily integrated.

In the domain of transportation, various ontologies can be developed based on the two dimensions explained above. However, in the context of transportation ontologies for smart cities, it is more effective to follow a formal ontology approach. Information or linguistic ontologies are appropriate when the goal is to prepare sketches or diagrams for a small project or when a glossary of words is required to define different concepts in the domain of mobility. In the context of smart cities, a more advanced ontology development approach is required that goes beyond just diagrams or glossary of words.

Regarding the scope and granularity of ontology, transportation ontologies have been generated at various levels. For example, some ontologies have been developed for specific functionality of Autonomous and Connected Vehicles (ACVs) (Viktorović et al., 2020), including the functionality of sensors in autonomous vehicles, and can be considered as application ontologies. Some other works define the core ontologies related to a specific mobility domain. For example, Zhao et al. (2015) developed a core ontology for Safe Autonomous Driving, that consists of minimum concepts and relationships between them for designing a safe autonomous driving system.

Moreover, some other ontologies have been designed for a specific domain of transportation, such as freight transportation or city logistics Anand et al. (2014), and can be categorized as domain ontologies. Finally, there are general transportation ontologies (Katsumi and Fox, 2017b; Bellini et al., 2014) consisting of several sub-ontologies describing a large transportation knowledge base.

In the context of smart city and smart mobility, ontologies have mainly followed a general formal approach. The reason is that the smart mobility domain consists of a huge number of entities, concepts, and the relationships between them. Hence, any ontology regarding smart mobility is usually an integration of several ontologies, sometimes from other domains of knowledge. The next section explains the most well-known ontology design approaches.

## 4 Ontology Developing Approaches

While various approaches have been used in the literature to develop ontologies (Falquet et al., 2011), to the best of our knowledge, an all-purpose methodology for ontology development does not exist. Indeed, no one can claim that there is a unique way to define the ontology of a domain. The best method to develop an ontology highly depends on the domain and application of ontology, as well as the extent to which a domain overlaps with other fields of science. However, this section explains some general issues worth considering when developing an ontology. Also, different design approaches are categorized in this section.

Ontology development approaches can be categorized based on two criteria (Falquet et al., 2011):

- Direction of taxonomy hierarchy
- Source type of ontology

Each of the criteria has been explained briefly in the following sections.

## 4.1 Direction of taxonomy hierarchy

The direction of taxonomy hierarchy describes three different approaches to develop class hierarchy (Noy et al., 2001; Hogan, 2020a):

- Top-down development process
- Bottom-up development process
- Hybrid development process

The top-down development approaches begin with the definition of the most general concepts (classes) in a domain, and subsequently, the sub-classes are defined. For example, we can start with the class vehicle, and then sub-classes can be defined as motorized and unmotorized vehicles.

Bottom-top approaches initially focus on defining the most specific classes and grouping the sub-classes to generate super-classes (Falquet et al., 2011). For instance, the car, bus, motorcycle, bike, and scooter are defined first, and then the car, bus, and motorcycle are grouped into class *motorized*, similarly bike and scooter as *unmotorized* class. Finally, a class *vehicle* is defined as the super-class of both *motorized* and *unmotorized* classes.

The hybrid approach combines both top-down and bottom-up approaches (Hogan, 2020a). For example, one begins with defining two middle-level classes, such as *unmotorized* and *motorized* vehicles, and subsequently define more specific sub-classes within each, i.e., car, bus, bike, etc., and finally add a super-class *vehicle* to the ontology. Also, one can generate more mid-level classes, such as *passenger* and *cargo* vehicles as the sub-classes of class *vehicle*.

None of these three methods can be considered superior or possesses an inherent quality over the others. Indeed, the preference among them is a matter of domain expertise and how an individual would rather view and define a domain. However, as the concepts in the mid-level are usually the most descriptive ones, researchers tend to pick the hybrid approach to begin the ontology development (Hogan, 2020a).

## 4.2 Source type of ontology

While the ontology development in some domains can be achieved via domain expertise and knowledge, identifying and defining the classes in vast and intricate domains is a time-consuming, high-cost, and contentious task (Falquet et al., 2011). Hence, researchers have used techniques to elicit knowledge from different resources, such as corpora, thesaurus, or relational database.

Such knowledge elicitation has led to the rise of learning methods in ontology development. The learning methods in ontology learning are mainly Natural Language Processing (NLP), text mining and Information Retrieval methods (Falquet et al., 2011) to elicit concepts from text documents.

The resources for knowledge elicitation can be categorized into different types:

- Text documents
- Schemata

Each category is explained in the following sections.

#### *4.2.1 Text documents*

Text documents include corpora, dictionaries, and thesauruses. Corpora are large and structured sets of texts on a specific domain that contain the concepts in the domain. Ontology learning based on corpora includes several steps, such as elicitation of the relevant terminology, identification of synonym terms, the establishment of concepts, organizing the hierarchical of the concepts, and inferring relationships between concepts and their properties (Falquet et al., 2011).

Thesaurus refers to a lexical ontology containing the definition of domain terms in a manner that the concepts, the relationships (for example, the synonyms of main terms, the more general and more specific terms) are explicitly defined. Thesaurus is mainly used to generate the first draft of a formal or software ontology (Falquet et al., 2011). Dictionaries, particularly the domain dictionaries, are also suitable for extracting the domain concepts and their relationships. All of these text documents can be analyzed by learning algorithms to extract the classes, relationships, and properties of a domain ontology.

#### *4.2.2 Schemata*

The second main source for ontology learning is Schemata, which are relational database models, Entity-Relationship (ER) diagram, Unified Modeling Language (UML), object-oriented models, or unstructured data (such as XML documents) (Hogan, 2020b; Falquet et al., 2011). Ontologies elicited from schemata are usually referred to as model-driven ontologies, as the ontology has been developed based on a previously available data model. Researchers have investigated different methods to convert schemata to ontologies. Such methods consist of a set of rules to map the structure of a source schema to an ontology, usually refers to schema mapping techniques.

Due to the structured form of relational databases, ontology development can benefit more from them than the unstructured text documents. The structure of the database is not only useful for defining the concepts and classes of ontology, but the data stored in the relational database is also beneficial for defining the class hierarchy and properties (Falquet et al., 2011). The main difference between relational databases and ontologies is that the latter has an object oriented schema, while relational databases follow a structured schema. Hence, while using relational databases for ontology development, the first step is to design a set of rules to transform the schema of relational databases into an object-oriented schema.

As mentioned above, construct ontology from the database usually follows certain mapping rules and principles (Desmond Mogotlane and Fonou-Dombeu, 2016). Some tools have also been designed to automatically construct ontology from relations databases, although the quality of resulted ontology may not be satisfactory always. Desmond Mogotlane and Fonou-Dombeu (2016) has reviewed the tools and algorithms to convert databases to ontology and compared them.

While ontologies have been primarily developed within the Artificial Intelligence community, the UML diagrams have been mainly utilized by the software engineering community as a standard language to visually represent a software system and its main actors, concepts, and the relations between them. Ontology and UML share similarities in many aspects. The main resemblance between UML and ontology is that both focus on objects and their properties; in other words, they are both following an object-oriented paradigm (Mejhed Mkhinini et al., 2020). Moreover, both UML and ontology first define the classes (concepts) and their properties, and afterward, they create instances of classes. Hence, there is great potential in ontology development using UML, and the field has attracted attention recently. Mejhed Mkhinini et al. (2020) have done a comprehensive literature review on the methods and tools to transform UML to OWL ontology.

Ontology can also be derived from different file formats, such as geospatial Shapefile, CSV, or XML files. W3C [1] organization (Berners-Lee, 2007) listed converters developed to convert from different file formats (around 44 different file formats at the time of this writing) to RDF (Resource Description Framework), that is a language for describing concepts and resources on the Web (RDF is explained in Section 5).

## 5 Ontology Languages

The previous section explained the main approaches for ontology design. Despite the availability of many methodologies and languages for ontology development, it is not straightforward to choose a proper method or tool for developing an ontology (Katsumi and Fox, 2017b; Kalibatiene and Vasilecas, 2011). Moreover, some methodologies have been developed based on specific ontology languages. Also, some ontology languages are highly dependent on a specific ontology development tool. Such a situation leads to the fact that there are not too many practical options to select for ontology development. Moreover, every ontology language or tool should be weighed up against its pros and cons to see if it is an appropriate tool or language for the target

[1] W3C is the World Wide Web Consortium and provides the main international standards organization for the World Wide Web.

domain application (Kalibatiene and Vasilecas, 2011; Hogan, 2020b; Katsumi and Fox, 2017b).

Ontology languages are the languages mainly developed in computer science used to construct ontologies and encode the knowledge in a specific domain and apply reasoning rules to process that knowledge (Smith, 1998). While there are many ontology development methods and languages, it is not easy to choose the appropriate language, mainly due to the fact that many of the existing ontology languages have been built for domain-specific ontology development. Moreover, developing ontologies requires an ontology development tool (Kalibatiene and Vasilecas, 2011; Bellini et al., 2014). However, deploying of some methodologies is restricted to certain tools. Hence, there are limited options to select from, for ontology development methodology.

At the top level, the ontology languages are categorized into two categories (Kalibatiene and Vasilecas, 2011): traditional ontology and Web-based ontology languages. The traditional ontology languages are mainly developed based on *first-order logic*. The second category has the languages developed based on Web standards to facilitate the interchange of data on the Internet (Kalibatiene and Vasilecas, 2011; Falquet et al., 2011). However, some languages belong to both categories.

According to Kalibatiene and Vasilecas (2011), there are five main ontology languages used in the literature:

- RDF (Resource Description Framework)
- OWL (Web Ontology Language)
- KIF (Knowledge Interchange Format)
- OIL (Ontology Interchange Language)
- DAML (DARPA Agent Markup Language)

The early works on Web ontology languages started early 1990's, including KIF (Genesereth and Fikes, 1992), OIL (Fensel et al., 2001), and DAML+OIL (Mcguinness et al., 2002) languages. Work on OWL and RDF development began by the W3C in 2001. Since then, RDF and OWL have been widely used in mobility ontology development projects and other ontologies across different fields of science. As most transportation ontologies have been developed using either OWL or RDF, these two languages are briefly introduced in the following section. The reader may refer to Kalibatiene and Vasilecas (2011); **?**); Mcguinness et al. (2002); Genesereth and Fikes (1992), for more information on other ontology languages.

## 5.1 RDF Language

World Wide Web Consortium (W3C) designed and developed RDF as a standard data modeling language to define and describe resources on the Web. Resources here refer to any sort of entity in the real-world, abstract, and virtual entities (Hogan, 2020a). The resources are defined using a standard expression of *subject–predicate–object*, referred to as triple-based format. Figure 2 demonstrates an example of triple format as a sentence. RDF uses a similar triple format to describe the resources, as shown in Figure 3. RDF language has been mainly designed to be understood and read by computers in Extensible Markup Language (XML) (Kalibatiene and Vasilecas, 2011).

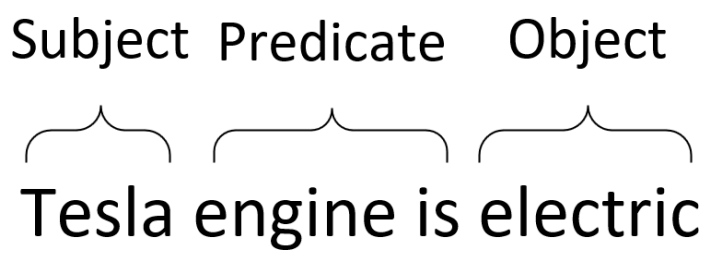

**Fig. 2** An example of subject-predicate-object sentence (image adopted from Hogan (2020a))

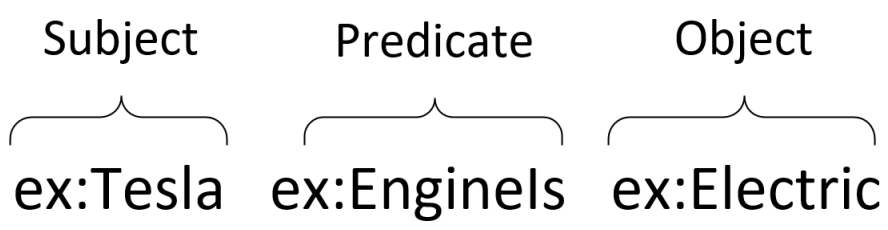

**Fig. 3** An example of subject-predicate-object in RDF language (image adopted from Hogan (2020b))

## 5.2 OWL Language

OWL is the most well-known Web-standard language for the semantic web, also developed by the World Wide Web Consortium (W3C). Like RDF, OWL has been used to formally classify comprehensive and complicated knowledge about concepts and entities, groups of entities, and relations between them. W3C has introduced OWL as a logic-based language designed to express knowledge so that computer programs can exploit the knowledge on the Web. Indeed, OWL can be considered as an extension of RDF language enhanced with more core vocabulary to provide a more comprehensive range of new terms (Hogan, 2020b). Hence, today, OWL is considered a much richer language to define the semantic web, making it a more appropriate language to integrate data from different sources automatically (Hogan, 2020b).

In the mobility domain, most ontologies have been developed in OWL language (Katsumi and Fox, 2017b; Bellini et al., 2014; Codescu et al., 2011), since OWL plays the role of a de facto language in the Semantic Web domain. Moreover, the majority of ontology software and editors have been developed using the OWL language. Hence, it is more straightforward for an ontology developer to choose among software designed for OWL language, rather than searching for specific software developed for other languages in a particular domain.

## 6 Ontology Design Procedure

While there are several approaches for the design and development of ontologies, some general rules and steps have been suggested in the literature to make ontology development an efficient procedure.

Before explaining the steps required to develop a mobility ontology, some general considerations should be noted (Noy et al., 2000; Falquet et al., 2011; Lorenz et al., 2005). First, generally speaking, ontology is inherently an iterative procedure and should be revised and updated regularly. Notably, as the transportation domain keeps changing fast due to technological advances and innovations, the ontology development should be considered an ongoing project continuously updated to reflect the new concepts and entities introduced by emerging technologies. Ontology development usually begins with an initial version, further extended and evaluated by domain experts against its pros and cons. For example, in the mobility domain, many disruptive technologies, such as Connected and Autonomous Vehicles (CAVs), will gradually set foot in the urban spaces in upcoming years. Accordingly, any mobility ontology should be revised to incorporate such new technologies.

Second, it would be easier for an ontology developer to consider the classes and relationships as nouns (classes) and verbs (relationships) in a sentence that describes some concepts in a domain. Figures 2 and 3, show how an example sentence describing an entity in the domain of mobility converts to classes and relationship in an ontology language.

While there are many approaches to design ontology, as explained in the previous section, one can benefit from a set of general-purpose rules applicable to ontology development regardless of the domain and data source. Noy et al. (2001) proposed a set of rules applicable for various fields of science, based on Protégé [2] software. The procedure is well suited for mobility ontology development either through domain expertise or by applying learning methods. Mobility ontology development may include the following main steps (Noy et al., 2001):

- Step 1. Identify the scope of mobility ontology and list important concepts and terms in the domain of mobility.
- Step 2. Consider reusing the existing mobility ontologies
- Step 3. Choose the direction of ontology development (as explained in Section 4.1), define the classes of mobility ontology and the class hierarchy.
- Step 4. Specify the properties of concepts (classes)
- Step 5. Determine the facets of the properties of each defined class
- Step 6. Generate instances of mobility ontology entities

The first step plays a crucial role in mobility ontology development, as the domain is vast and multidisciplinary. For example, for *vehicle* ontology, the scope and context in which the ontology will be developed largely impact the next steps in the development of mobility ontology. As an example, a car manufacturer views a *vehicle* with more details than a researcher who aims to develop *vehicle* ontology in the context of smart mobility. A car manufacturer may be involved with the detailed mechanical and electrical aspects of a vehicle or its performance and appearance, while such properties of a vehicle may not be in the focus of a transportation engineering researcher.

[2] Protégé is an open-source ontology editor, which is indeed a platform with a suit of tools to develop ontology and visualize it. Protégé also contains a rich library of ontology across different fields of science.

After identifying the scope of mobility ontology, the researcher needs to list all the concepts and terms in the domain. For a mobility ontology, all the Concepts, such as streets, intersections, vehicles, walkways, traffic devices, etc. should be enumerated. Also, detailed relations between the concepts should be listed. This step is important, as the classes and properties will be further defined based on the concepts listed in this step.

Step 2 recommends to search for the available ontologies in the literature and consider reusing them. Especially in a vast domain like transportation, some ontologies already developed across different domains that one can benefit from. For example, Linked Open Vocabularies for the Internet of Things (LOV4IoT) Website consists of more than 550 ontologies (Noura et al., 2019), including smart city and mobility ontologies, at the time of writing. Protégé Ontology Library is another source of developed ontologies across different science fields and a good source to find the available ontologies and consider reusing them.

However, although the idea of ontology reusing may sound appealing, it is not straightforward to implement due to several reasons. Obrst et al. (2014) have discussed "Mismatches and Misunderstandings", "Finding The Right Ontology" and "Integration" as the major hurdles in reusing an existing ontology. Mismatches between the domain coverage and requirements are caused when the domain scopes are entirely different or while the ontology developers look at the same entities from different perspectives. Sometimes an available ontology is very specific, such as an ontology developed for defining the autonomous driving decision-making procedures, and may not be straightforwardly reusable in a broader context, such as including autonomous vehicles ontology in a smart city ontology. Some existing ontologies may have classes and properties utterly irrelevant to other domain concepts and hard to integrate. If not overcome, such hurdles may finally discourage ontology developers from reusing existing ontologies, and instead, they may decide to develop it from scratch. Furthermore, ontology reuse usually requires integrating several existing ontologies, which sometimes may be more challenging than developing a new ontology. Moreover, although there are mobility ontologies available, some of them are not well documented or reusable (Katsumi and Fox, 2017b). Despite these facts, standard foundational ontologies, such as time or geospatial ontology (explained in Section 7.1) have been developed and well documented by some organizations, such as World Wide Web Consortium (W3C) and are used widely across different domains.

Step 3 and 4 are the most important steps as the classes and properties are defined, and the relationships between them are specified. The class hierarchy is also defined in step 4, based on the approaches explained in Section 4.1.

Classes can include four types of properties: (Noy et al., 2001; Hogan, 2020b):

- intrinsic properties such as ”number of doors” or ”fuel type” of a vehicle.
- extrinsic properties such as vehicle ”brand” or ”production date”
- parts can be physical or abstract parts. For example, different parts of a road network, such as road links and intersections.
- relationships define the relations to other entities. For example, a road link *isConnectedTo* an intersection.

Another issue regarding classes hierarchies is the notion of sub-class and super-class. Any class can have one or more super-classes. While a class has more than one super-class, the super-classes can be mutually exclusive. For example, the owner of a vehicle can be an individual or an organization. Hence, one can define a class *owner*, which is a sub-class of two super-classes: class *organization* and class *individual*. However, class *owner* can have only one super-classes, either *organization* or *individual*. It should be mentioned that any subclass of a class inherits all the properties of that class.

In step 4, the facets of properties are defined. Facet definition refers to three issues: the values and data types of properties, as well as the range and domain of a property. For example, the class *vehicle* can have *make* as a property. The property *make* is of data type *String* and accepts the make of the vehicle as values, and its range is all the available make of vehicles in the market.

Occasionally, the value of a property can be the instance of another class. For example, in transportation ontology, one can define two classes: *vehicle* and *organization*. Class *Organization* refers to a group of individuals who work together towards a goal. Class *vehicle has* a *Manufacturer*, and its values can be the instances of class *organization*.

The last step in ontology development is creating instances of classes in the hierarchy. Individual instances of a class are first created, and then the properties of the class are filling out by their values. For example, for the class *vehicle*, one can create an instance *Tesla*, to represent a specific vehicle. This instance can have the following properties:

- *Manufacturer* : Tesla, Inc.
- *Fuel Type*: Electricity
- *Number of doors*: 4
- *Owner* : Peter

Figure 4 demonstrates an example of the vehicle, organization, person ontology and the properties of class vehicle. The organization ontology has been explained in Section 7.1.7.

## 7 Mobility Ontologies

Smart mobility includes many different dimensions and sectors. It also overlaps with other domains such as urban and land-use planning, Internet of

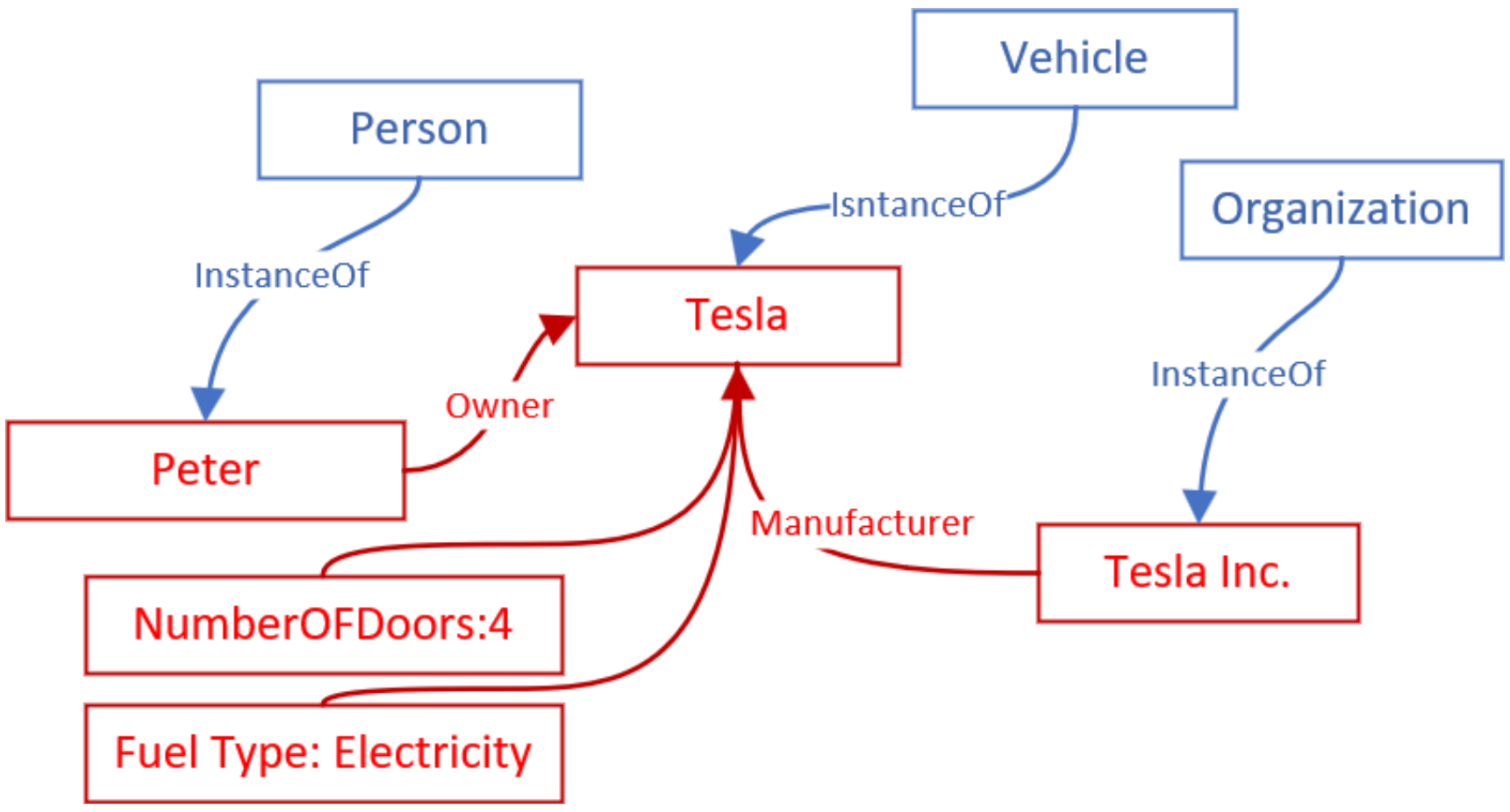

**Fig. 4** An example of vehicle ontology, with classes shown in blue and properties and instances in red.

Things, Sensors, etc. To show various aspects of smart mobility and the ex- tent of the domain concepts, two scientific platforms, i.e. Web of Science and Scopus were searched for "smart mobility" phrase. Figure 5 shows various fields of science, with the most number of publications on "smart mobility", based on Web of Science data analysis. The inclusion and exclusion criteria for selecting publications and analyzing their data are presented in Table 7. The search resulted in 1,779 and 1,812 publications from Scopus and Web of Science, respectively. Smart mobility itself includes many parts and sectors, as shown in Figure 5. The image shows the co-occurrences of keywords in Web of Science publications on smart mobility. While there are numerous publications on smart mobility, they are much less when searching for "smart mobility" and "ontology" in Web of Science or Scopus databases. Keeping the same inclusion criteria in Table 7, there were less than ten publications in each database, which shows the lack of studies regarding the ontology of smart mobility in the current literature. However, one should notice that not all the available ontologies on transportation or mobility are presented in conference or peer-review papers. Indeed some of them have been published in the form of metadata data models using OWL or RDF frameworks. Moreover, some ontologies have been published in other domains related to smart cities, such as connected and autonomous vehicles, that are not included in the search criteria.

The transportation system can be categorized into two general categories: transportation physical network and transportation cyber network. Each of these categories includes several foundation or domain ontologies. The transportation physical network has been explained in Section 7.2. The foundation ontology (which are generic top-level ontologies used to explain basic concepts in a domain ontology, and usually are re-used from previously developed ontologies) are explained in Section 7.1. Due to innovative and advanced tech-nologies in cyber network ontologies, a separate section, i.e., Section8, is devoted to explaining the cyber network and future directions in transportation ontology in the concept of the smart city.

| | **Inclusion Criteria** | |
|---|---|---|
| **Database** | Scopus | Web of Science |
| **Publication period** | 1984-2021 | 1991-2020 |
| **Document type** | Article, Book/Book Chapter, Review, Editorial | |
| **Source type** | Journal, Book/Book Series, Conference Proceeding | |
| **Subject Area** | Engineering, Computer Science | Shown in Figure 1 |
| Language | English | |
| Number of documents | 1,779 | 1,812 |

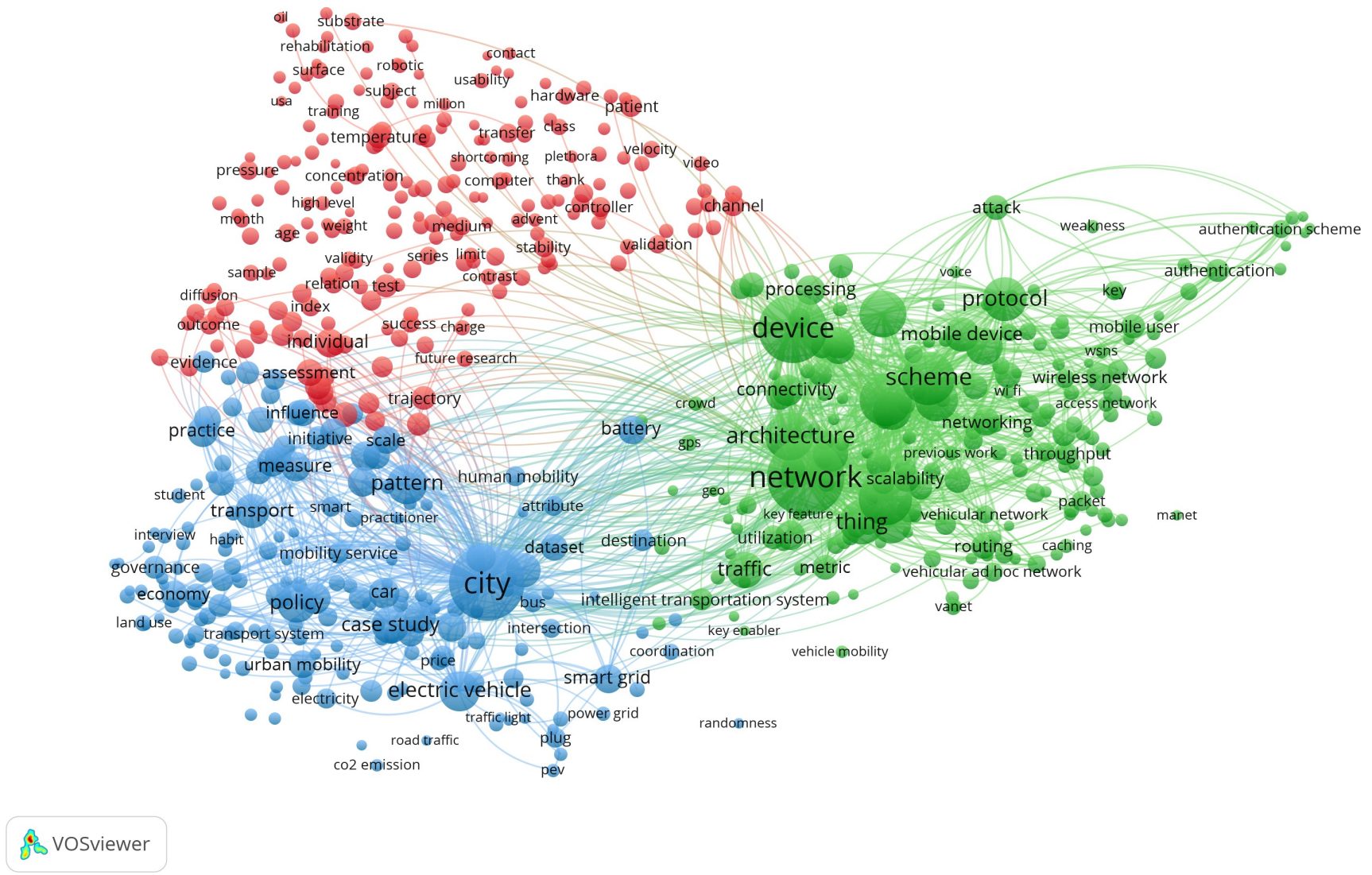

**Fig. 5** Most Frequent Words in "Smart Mobility" Publications on Web of Science database.

## 7.1 Foundation ontologies

Foundation ontologies cover the basic ontologies used across different domains. For example, the ontology of time and space has been initially developed in Geography domain, however, such ontology has been re-used by another domain studies, such as transportation or urban planning. Foundation ontologies can be categorized into the following basic ontologies as suggested by (Katsumi and Fox, 2017a):

- Geospatial
- Time

- Weather
- Trip
- Units of Measure
- Change
- Household and Dwelling
- Organization Ontology
- Stakeholders (public/private actors)

In the following, each of these ontologies is explained, and best practices to reuse are mentioned.

### *7.1.1 Geospatial Ontology*

Every entity in the smart city or smart mobility domain requires a spatial property to be located on a map. An intersection or a street cannot be shown on a map without their geospatial information. The location of entities can be presented in different forms:

- Longitude and Latitude Coordinates
- Point
- Linestring
- Polygon

Longitude and latitude coordinates are designed to point to any location on the earth's surface and widely used in many GPS positioning devices. Point is mainly used for showing the location of vehicles, individuals, or dwelling. Linestring is used to show any line shape entity (either a straight or curved line), such as road network links or railways. Polygon is usually used to presents the boundaries of a city, a municipality, a transit agency, etc. It also can be used as the shape of a vehicle in self-driving cars, where the geometric shape and boundaries of a vehicle play an important role in designing the autonomous driving algorithms.

For example, in iCity ontology (Katsumi and Fox, 2017b), every entity has a spatial feature. The spatial feature has been defined by reusing the WGS-84 Ontology, defined by W3C and Linked Open Vocabulary as in RDF format.

### *7.1.2 Time Ontology*

Time Ontology defines the required classes for describing the temporal properties of real-world assets or resources on the Web. The most well-known Time ontology is the "Time Ontology in OWL" (Hobbs and Pan, 2006) developed by W3C, which is widely included in many ontologies across various domains. Time has also been encoded into upper-level or domain-specific ontologies using RDF languages (Gutierrez et al., 2006). To equip ontologies with Time, one should take considerations to avoid duplication of the original ontology, specifically when the entities of an ontology have properties that change over time. (Krieger, 2010) has proposed a general methodology to equip ontologies with Time ontology. The transportation ontology can take advantage of the proposed methodology as the properties of many transportation entities, such as vehicle or passenger GPS location, are changing over time. Today, many ontologies in OWL use the Time ontology developed by W3C.

### *7.1.3 Weather Ontology*

Transportation systems are significantly affected by weather conditions. Road weather management decisions require access to data on environmental conditions from observing systems and forecast providers. Moreover, the adverse effect of weather conditions on traffic accidents is a well-known phenomenon in traffic accident analysis and prevention studies. Furthermore, weather conditions play an important role in the functionality of autonomous vehicles. Indeed, to conceptualize the environment in which an autonomous vehicle evolves, the weather ontology is essential. Chen and Kloul (2018) developed weather ontology for advanced driver assistant in OWL language. A more sophisticated weather ontology has been developed by KM4City (Bellini et al., 2014) in the context of smart cities. They have developed the weather ontology based on the real-time weather forests. The ontology has been published in RDF format using SPARQL query language.

Weather ontology may consist of a different number of classes or slots depending on the specific domain requirement. For example, the weather ontology for aviation and air transportation is more comprehensive and detailed compared to the ontology used for road transport. Moreover, in road transportation and traffic monitoring systems, the weather ontology can possess different classes and slots based on data types collected by different meteorological instruments and sensors.

### *7.1.4 Units of Measure*

The Ontology of units of measure refers to a set of classes defined to describe the different measures or values of a given quantity (Katsumi and Fox, 2017a). For example, a vehicle with 4 meters length can have a sub-class called *length*, which describes the nature of the length quantity, its numerical value (i.e. 4), and the unit that length is measure in, which is meter in this example. In this case, instead of defining length as a property of the vehicle, defining length as a sub-class of vehicle provides a more comprehensive description of length. Moreover, the class *length* can be used to describe the length of any other entities.

### *7.1.5 Change*

Change ontology firstly proposed by Welty et al. (2006) to include fluent properties of concepts in OWL. In mobility domain, several concepts have changing properties. For example, the location of a vehicle is a changing property (Katsumi and Fox, 2017a). Other example is the number of passengers on a shuttle at different times of a day. Even the transportation network itself changes over time, for example, the number of traffic lanes or pedestrian lane width may vary over time. Change ontology indeed enriches an ontology by adding a fourth dimension to the concepts and entities which have a changing property (Welty et al., 2006). In the domain of transport, the iCity ontology (Katsumi and Fox, 2017b,a) has included change ontology in their general ontology of transport. Indeed, they have followed the change ontology suggested by Krieger (2008) and adopted it for the domain of mobility.

Change ontology also makes some complex data inferences possible and ensures that some competency queries are correctly answered. For example, if the result of a query is to answer whether a vehicle has taken a trip, it can query about the change property of the vehicle location to see if it has changed over a period of time.

### *7.1.6 Household and Dwelling Ontology*

In transportation planning, demand modeling is usually conducted for individuals in a households, as the household members usually depend on each other for commuting, as they usually share a car, also they adopt similar mobility habits. A household also occupies a dwelling unit. In transportation ontology, a dwelling may have a market value, a location or address (Katsumi and Fox, 2017b), and type of dwelling. Other detailed properties of a dwelling, such as the building sensors, are not in the scope of transportation ontology.

### *7.1.7 Organization Ontology*

An Organization usually refers to a body of individuals, in private or public sector, who follow the same goal(s) (Katsumi and Fox, 2017a). Organization ontology, like the stakeholder ontology, has been developed mainly in e-Government and project management ontologies.

### *7.1.8 Stakeholder Ontology*

In the field of transportation, the concept of "stakeholder" is mainly used in freight transportation studies or in transportation externalities assessment, i.e. the study of negative impacts of transportation activities on other sectors of society. Stakeholder ontology in logistics and freight transportation is described in Section 7.2.5.

### *7.1.9 Trip*

Trip describes the mobility of persons or goods from an origin to a destination via a transport mode. In the transportation domain, trip ontology has been developed within iCity ontology (Katsumi and Fox, 2017b). Trip ontology has also been studied in other domains, such as Recommendation Systems (Choi et al., 2006). However, the class *trip* can consist of different *trip segments*. Also, the trip itself can be considered as part of a *tour*. Hence, the trip definition in mobility ontology highly depends on the domain application, i.e., whether the domain experts aim to consider tours in their ontology.

## 7.2 Transportation Physical Network ontologies

Mobility ontology can be categorized into several sectors. However, at the top-level mobility sector can be categorized as Physical and Cyber Network. Physical networks itself includes different types of transportation concepts:

- Road Transportation Network
- Pedestrian Network
- Cycling Network
- Railway Network
- Transit System
- Freight Transportation System
- Road Service Area and Parking
- Vehicle

The Cyber network in the transportation domain consists of different innovative and disruptive technologies, such as sensors and Internet of Vehicles (IoV) communications. Moreover, some innovative technologies, such as autonomous robotics, Connected and Autonomous Cars (CAVs), or Unmanned Aerial Vehicles (UAVs), will flood into smart cities in succeeding years. Hence, Section 8, Smart Mobility and Future Directions, has been devoted to four major innovative technologies:

- Sensors
- Autonomous Robotics
- Mobility as a Service (MaaS)
- Connected Roadways and Internet of Vehicles

The next sections explain each of the above-mentioned ontologies.

### *7.2.1 Road Transportation Network*

Road network ontology is among the essential ontologies in every transportation ontology development. The road transportation network consists of links (in some studies (Lorenz et al., 2005) referred to as edges) and nodes. The road links are streets and roads on which motorized vehicles move. Nodes represent the connections between the links in the transportation network. Nodes can be located on intersections or in the middle of a street when the characteristics of the street, such as speed limit or the number of lanes, change. One of the best practices of road transportation ontology has been implemented by INSPIRE (INS, 2014).

The INSPIRE transportation network ontology scope is vast and includes five major transportation networks: Road, Railway, Water, Air transportation, and Cableways. The INSPIRE ontology defines the details of each network and the connections between them.

Road network ontology has also been developed by The Ontology for Transportation Networks (OTN) (Lorenz et al., 2005). Both INSPIRE and OTN ontologies have been developed as part of a geospatial information project, and the data specification in both is based on a common template used for different domains, such as transport, geoscience, etc. However, OTN ontology has not been updated since 2005, the date it has been published first.

Another ontology developed in the smart cities domain is the km4City ontology (Bellini et al., 2014), which includes road network ontology. The Km4city ontology has defined several entities regarding road transportation network, such as Road, Node, RoadElement, StreetNumber, RoadLink, Junction, Entry, and EntryRule Maneuver.

All the above-mentioned road network ontologies are comprehensive and include all the details related to the road network. Moreover, all of them have mainly generated from the encoding of Geographic Data Files (GDF) into OWL language.

A road network can possess different classes. However, the simplest road network consists of road link sequences, road links, and nodes, as shown in Figure 6. In this figure, the classes are presented by the blue boxes and properties by the white ones. A road link sequence consists of road links, usually have the same properties or are part of a route. Figure 6 represents only a few numbers of properties. Apparently, road links, for example, can have other properties, such as pavement marking, capacity per direction, right of way, etc.

### *7.2.2 Railway Transportation Network*

Railway transportation network includes railway links and nodes, and railway yards and their properties. It should contain all types of railways for both passenger and freight transportation. All the mobility ontologies developed in the previous section have defined railway ontology, such as INSPIRE, iCity, OTN, and Km4City.

However, Network Statement Checker Ontology (NSO) (Verstichel et al., 2011) is another ontology developed mainly for railway networks. The study's goal is to design a system to check the feasibility of running a train on any specific railway line. While the core of the NSO ontology is the railway network, the classes are defined as very generic and applicable to other transportation networks (Katsumi and Fox, 2017b). One of the distinctive features of the NSO ontology, compared to the other mobility ontologies, is that it has distinguished between physical tracks and transit lines. Their approach is first to define the railway network elements and then collect information about these elements provided by other transit and railway authorities. This approach is useful and more practical when the ontology is developed for a specific application, which is a railway network checker in the case of NSO ontology.

An example of classes and properties in railway transportation ontology has been shown in Figure 7. Again, it should be mentioned that this writing does not present all the possible classes in a domain ontology; instead, it provides just the most important classes in a category as examples. Railway Line, Link Sequence and Link, as well as Railway Node and Railway Yard Node are the classes in Figure 7. The Railway Yard Node is a sub-class of Railway Node, but with a restriction (shown as a red box in Figure 7) on the Form Of Node. Indeed, for a Railway Node, form of node can be defined as Junction, Level Crossing, Pseudo Node, Stop, etc. However, the Form of Node property for a Railway Yard Node is always Railway Stop.

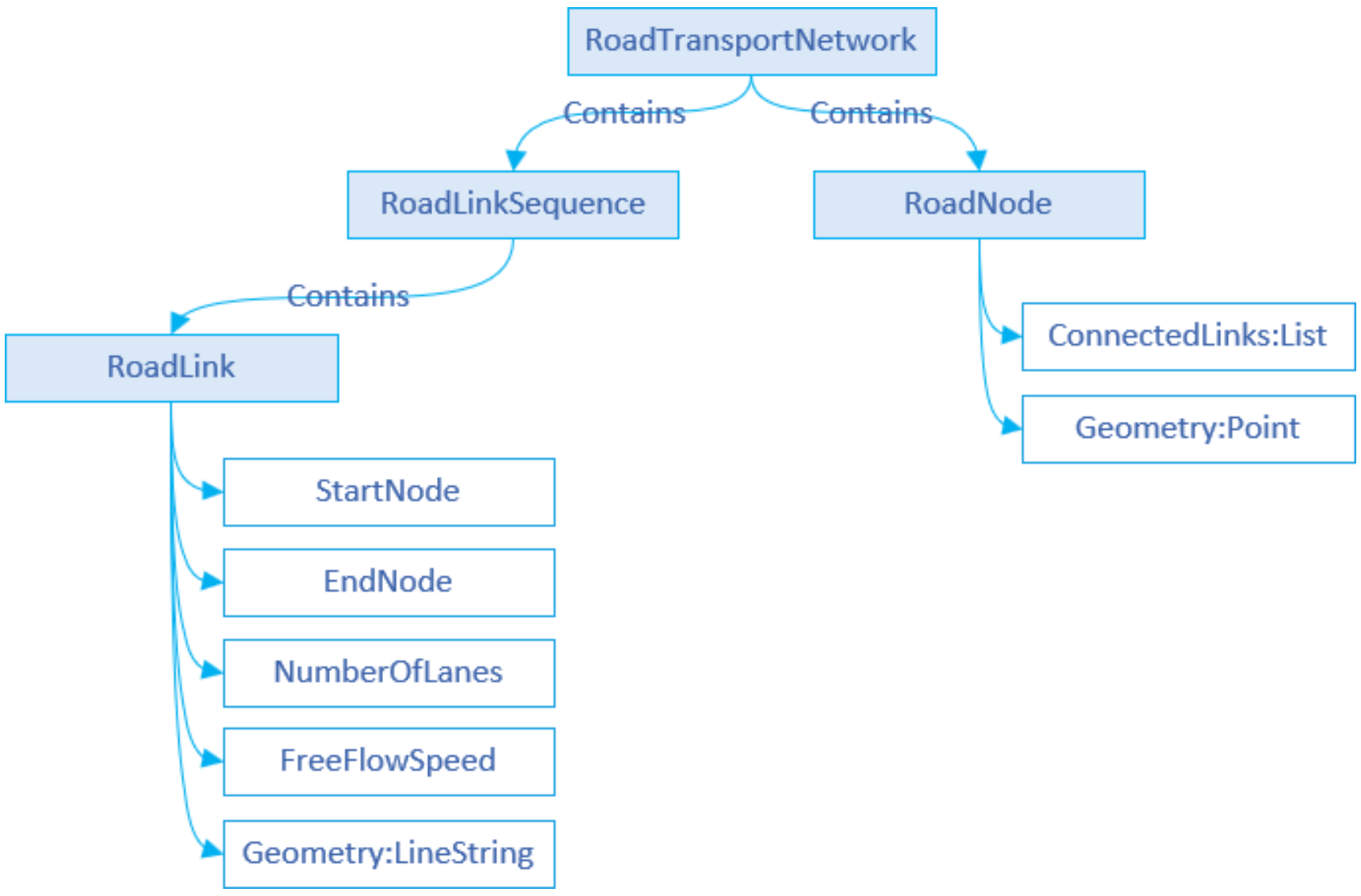

**Fig. 6** An example of classes and properties in a transportation road network.

### *7.2.3 Cycling and Pedestrian Network*

Inclusion of cycling and pedestrian networks in an ontology is achieved via two approaches: first, by including a “form of way” or “road type” property for road transportation network and distinguish motorized road segments from cycling and pedestrian segments. The second approach is to define an independent cycling network and make connections between the road network and cycling network wherever is required.

While many transportation onotologies (Katsumi and Fox, 2017b; Bellini et al., 2014) defined pedestrian and cycling network as part of the road transportation network, however, due to the unique characteristics of pedestrian network or cycling network, it is better to define them as separate entities. For example, parts of the pedestrian network in today’s cities are extended to underground and in-building walkways, distinctive from road networks where motorized vehicles can travel. Bike and pedestrian trails are usually located on areas not accessible to motorized vehicles and may not be included in the road network. Moreover, cities and government authorities usually publish geospatial data of pedestrian or cycling network independently from regular road network used by cars and buses.

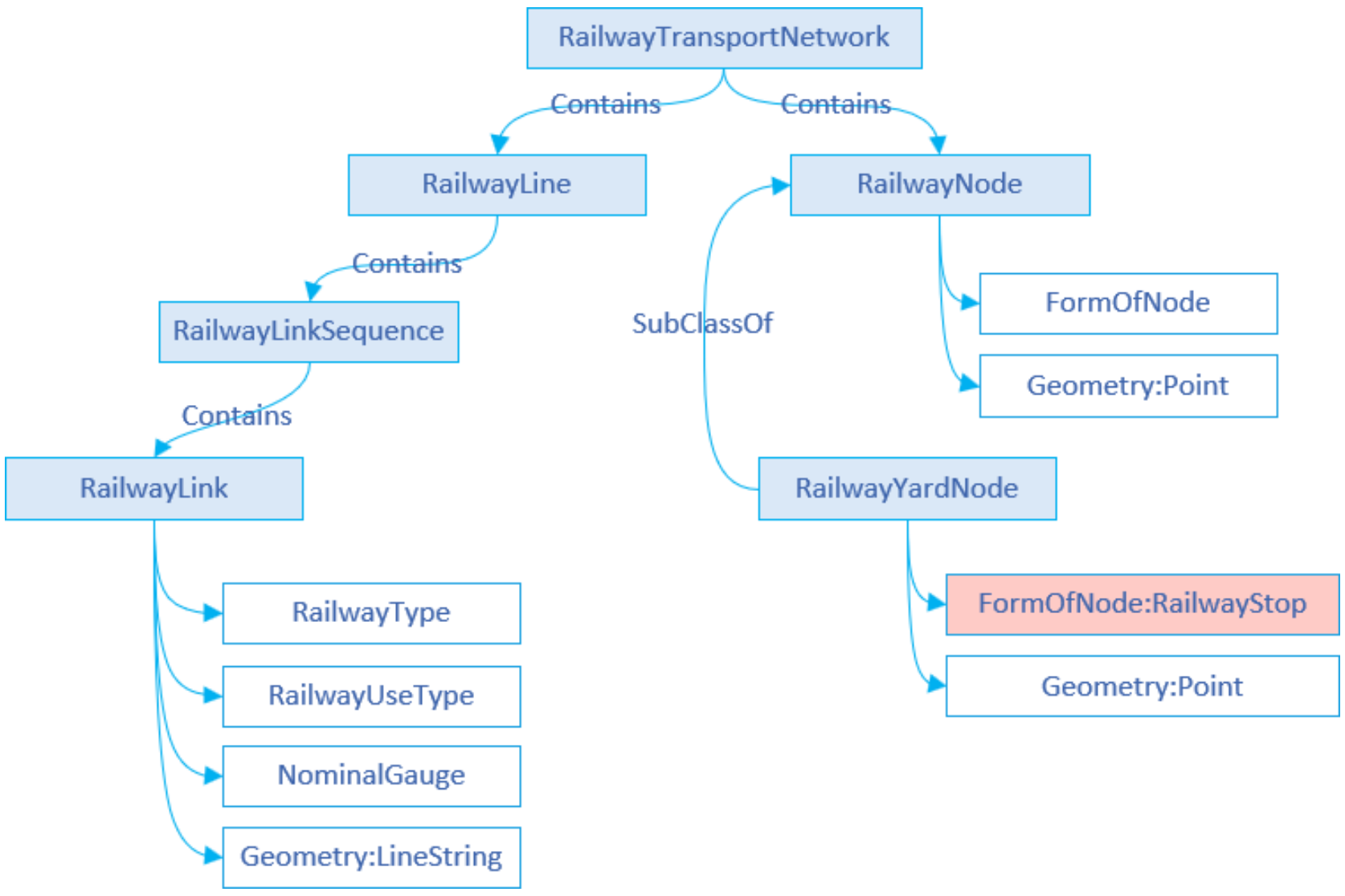

**Fig. 7** An example of classes and properties in a railway network.

### *7.2.4 Transit System*

The Transit network includes the bus and metro routes, their associated stops and schedule, and their geospatial information and access method, i.e., the payment method for accessing the public transit services. iCity ontology (Katsumi and Fox, 2017b), OTN ontology (Lorenz et al., 2005), as well as Km4City ontology (Bellini et al., 2014). They have similarly defined the main concepts in transit systems, as mentioned above. However, all of them lack considering static and real-time GTFS (General Transit Feed Specification) data (GTF, 2017) in their ontology. Indeed, GTFS data is considered as Linked Open Data and can enrich the transit system ontology by the inclusion of a detailed transit schedule and its geospatial features.

Figure 8 shows an example of classes, properties, and relationships for transit network ontology. Transit Route and Transit Stop are the main classes in transit ontology. One may also define a Transit Schedule class as the property of a transit route or node. The data for the transit schedule class can be acquired from GTFS data. Every transit system has a fare collection method that determines the transit access method (the values can be cash, transit pass or mobile app, etc.), the fare type (monthly pass, weekly pass, pre-purchased tickets, etc.), or validity period. Moreover, the transit system may include a transit shelter that kind of street furniture serves transit users.

Transport Disruption Ontology (TD) (Corsar et al., 2015) has also incorporated transit events in its ontology. The TDO ontology's focus is to design a data model to recognize events with a major impact on travel patterns. The ontology is used to query and extract social media data, in this case, any event reported on social media, and subsequently provide travel information to transit users.

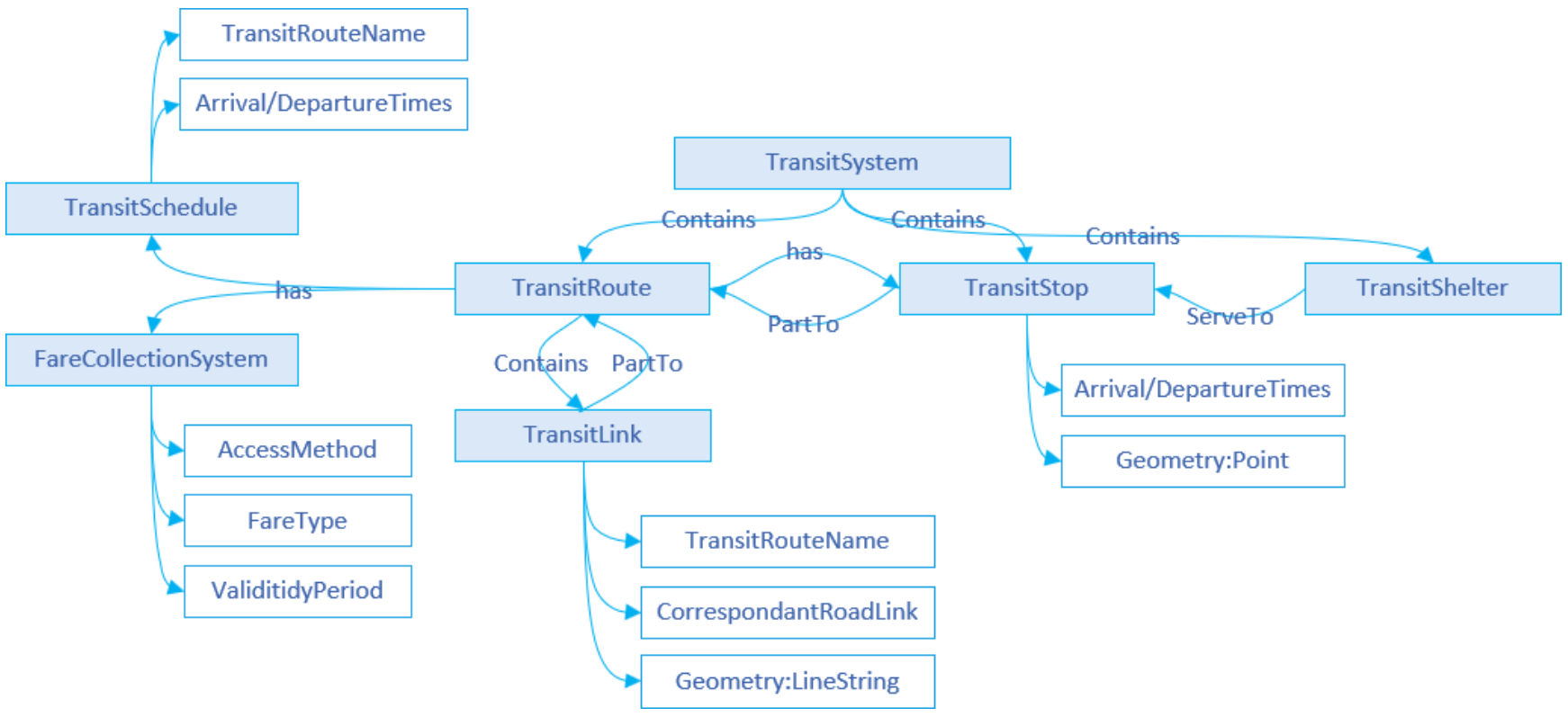

**Fig. 8** An example of classes and properties of a transit system.

### *7.2.5 Freight Transportation System*

The urban freight transportation system plays an important role in today's cities' logistics and economy. The most well-known ontology regarding freight transportation is GenCLOn ontology (Anand et al., 2014), which is an ontology for city logistics. Urban logistics is a domain dealing with freight transportation issues that different stakeholders in urban areas are faced with.

Different concepts (classes) should be included in the development of freight transportation ontology, such as warehouses, fleets, and products. Figure 9 shows an example of classes and their properties and relationship in an urban freight transportation system. The stakeholders in the domain of urban logistics can be categorized into two general classes as *Public Stakeholder* and *Private stakeholder*. Any *private stakeholder* is also a sub-class of either *organization* or *person* class. Here, we say class *organization* and class *person* are mutually exclusive super-classes of class *private stakeholder*. Also, class *private stakeholder* itself has three sub-classes as: *shipper, carrier and receiver*, defined by Anand et al. (2014).

The class *public stakeholder* is also classified into three sub-classes: *National, Regional and Local authorities* classes. Each stakeholder can possess several properties, for example, a stakeholder has an *objective* and a *location*. Also, each shipper and carrier may have a logistics fleet, mainly a list of vehicles used by a stakeholder to deliver products.

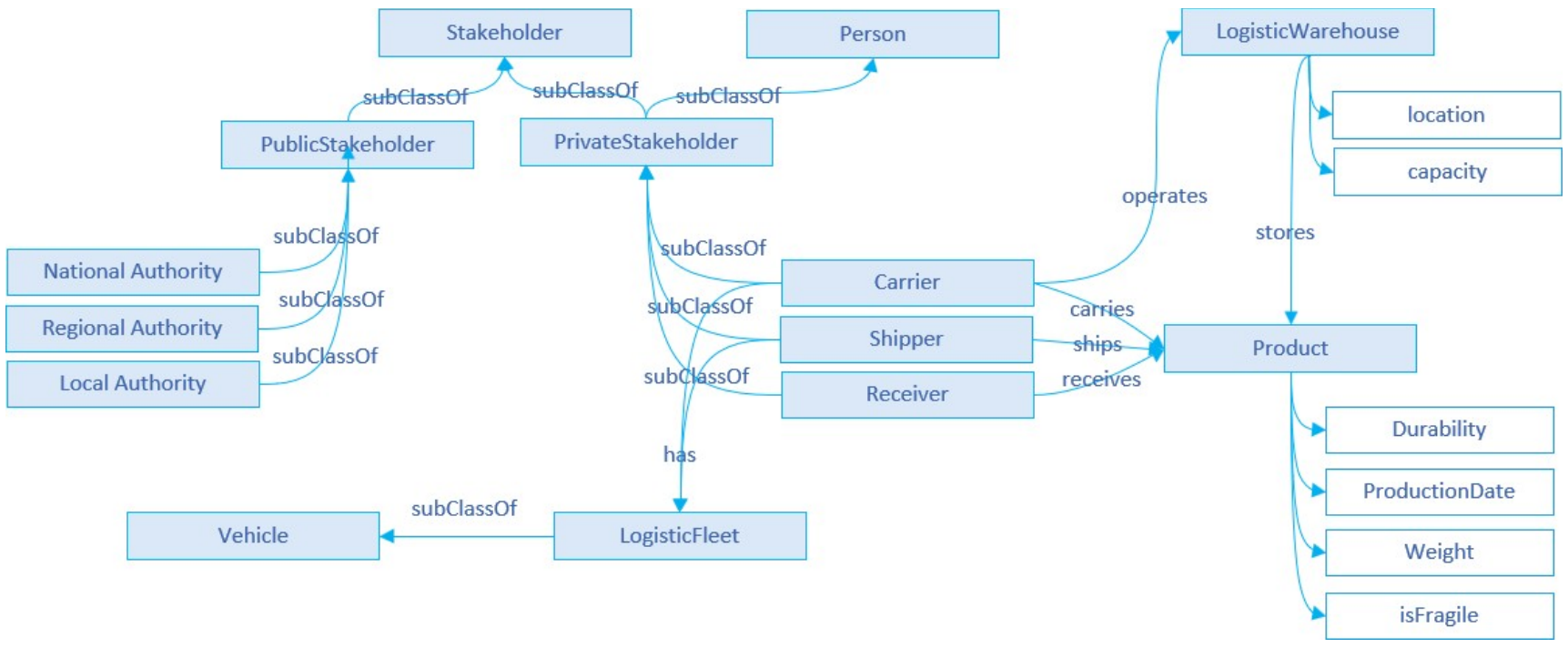

**Fig. 9** An example of classes and properties in urban freight transport.

While there are some ontologies developed in the field, in the foreseeable future, freight transportation will stand to benefit from innovative delivery methods, such as unmanned aerial vehicles (UAVs) for goods delivery. Moreover, UVAs application in goods delivery changes not only urban logistics, but also introduces new concepts to the urban mobility, such as human–robot interactions. They occur when robots require some kind of human intervention to function properly. For example, assigning a fleet of UAVs to different delivery tasks requires a robust system to efficiently manage the UAVs fleet by a human and maintain them in the case of a fault detection (Kumar et al., 2019). Such requirements of UAVs also will change the future of freight transportation ontology. Section 8 explains more on innovative autonomous robotics.

### *7.2.6 Road Service Area*

Road service area refers to any surface annexed to a road network that usually offers some services to road users. Some ontologies (INS, 2014) in the literature have defined road service area as an independent class, which can have different road service types, such as gas stations, rest and drive through area, toll area, or parking. However, the majority of current mobility ontologies lack the road service area related classes. For example, three ontologies, i.e. Ontology for Transportation Networks (OTN) (Lorenz et al., 2005), iCity ontology (Katsumi and Fox, 2017b) and Open Street Map Ontology (Osmonto) (Codescu et al., 2011), have only considered parking area in their ontology.

## 8 Future Directions in Smart Mobility Ontology

This section points to the cyber-physical transportation networks as well as innovative and disruptive technologies and services in the domain of mobility that have not been covered extensively by the current general ontologies in the literature. The innovative or disruptive transportation technologies and services can be categorized as the following areas:

- Mobility Sensors
- Mobility as a Service (MaaS)
- Autonomous Robotics
- Connected Roadways Technologies and Internet of Vehicles

All of these categories have been developed in recent years and are either currently available or forthcoming in the market. However, no ontology has been developed yet for these technologies. This section explains each of the above innovative areas. Furthermore, it introduces some ontologies developed in other fields of science related to each area above. Moreover, the main concepts or categories in each of the above areas are introduced. Ontology developers who aim to cover disruptive and innovative technologies in the smart mobility ontology will benefit from this chapter.

## 8.1 Mobility Sensors

Generally speaking, sensors are electronic devices that detect and react to changes in the sensor's environment. The change usually reveals the property value, and the process to determine this value is referred to as *observation* (Janowicz et al., 2019b). Observation usually includes the sampling process from the feature of interest, for example, a loop detector observes the presence of vehicles, and it takes *samples* from a specific area of a road. Observations can trigger some actions, called actuations, and the devices or entities that perform the actions are referred to as *actuators*.

Today, a variety of sensors collect data in smart cities, from Bluetooth or WiFi sensors to thermal and 3D cameras, which lead to the huge amount of sensors data published on the Web, which consequently has given rise to the reuse and fusion of such data (Janowicz et al., 2019b) in smart mobility stud- ies. However, such data are mainly published as raw data without any context or information required to interpret or analyze the data. Hence, to facilitate the integration and fusion of sensors data, a standard framework and linked vocabulary that include sensors, observations, samples, and actuators, is an essential part of many smart mobility studies. Currently, the most comprehensive and update ontology regarding sensors is Sensor, Observation, Sample, and Actuator ontology (SOSA) Janowicz et al. (2019a), jointly developed by W3C and OGC (Open Geospatial Consortium) Spatial Data on the Web (SDW) Working Group. SOSA is a flexible, comprehensive, and coherent ontology to define any entity, relationship, and activity (Janowicz et al., 2019b) regarding sensors, samples, and actuators. SOSA also has been designed as an extendable vocabulary easily being combined with other ontologies. Figure 10 shows an example of classes and properties in SOSA sensor ontology.

Regarding on-vehicle sensors, two ontologies have been developed based on SOSA. One is Vehicle Signals and Attribute Ontology (VSSo) (Klotz et al., 2018), which has been developed to understand and define the vehicle-specific signals, based on Vehicle Signal Specification (VSS) taxonomy. Another ontology called Connected Traffic Data Ontology (CTDO) has been developed based on the foundation of SOSA ontology. The focus of CTDO is on the on-vehicle sensors. On-vehicle sensors refer to any type of sensor installed on modern vehicles, such as cameras and radar sensors. There are several types of radar sensors on today's vehicles, like Blind-Spot Detection (BSD), Autonomous Cruise Control (ACC), Rear Cross Traffic Alert (RCTA), Rear Cross Assist (RCA), and Safer Lane Changes and Overtaking sensors **??**. It should be mentioned that a modern vehicle may possess several other sensors, many of them control the vehicle engine's performance, and probably are out of scope for smart city ontologies.

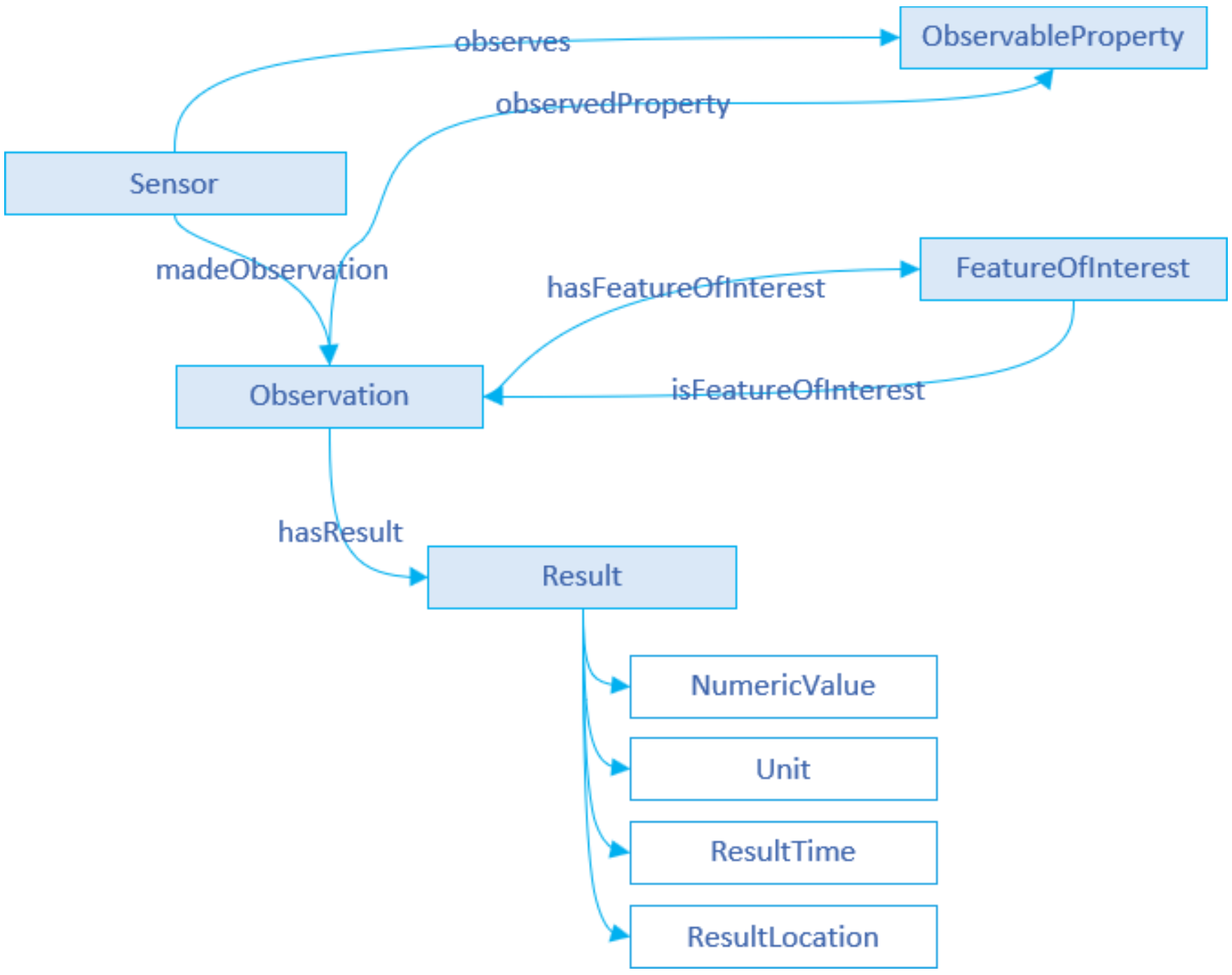

**Fig. 10** An example of classes and properties in SOSA sensor ontology.

While on-vehicle sensors play an important role in smart mobility and traffic safety, none of the currently available transportation ontologies have incorporated such sensors in their vehicle ontology. Furthermore, besides on-vehicle sensors, there are other two types of sensors in the smart mobility domain: intrusive on-road sensors, which are mounted on or under the road surface, and non-intrusive road-side or above-road sensors, which are sensors mounted on the poles at the side of the road or mounted above the road on sign bridges (Klein, 2017). Inductive loops and magnetic sensors are among the intrusive sensors, while video detection systems, Lidar (laser radar), passive infrared, and ultrasound sensors are categorized as non-intrusive sensors.

As mentioned above, sensors ontology has been developed mainly in the traffic management system domain or connected and autonomous driving. However, the current mobility ontologies, such as Km4City, OTN, iCity or OSMOnto lack the sensors ontology. In the context of the smart city, all the mobility-related sensors should be defined in the smart mobility ontology, and the relation between them, as well as the connections between on-vehicle sensors and road sensors, should be defined using the same ontology paradigm. Moreover, the current ontologies can be extended to integrate mobility sensors. For example, the current vehicle ontologies can be extended to include a new property representing all the safety and traffic sensors installed on a vehicle.

## 8.2 MaaS

MaaS refers to multiple mobility services, usually integrated into a Website or smartphone app that enables the individuals to plan, book, and pay based on their mobility needs and preference. Maas includes several innovative services, such as ride-hailing, car-pooling, or bike-sharing systems, to name but a few. Ontologies in the realm of transportation have not yet defined MaaS concepts, though, in recent years, there has been a major interest in Maas services.

In general, MaaS includes two major parts: fleet and infrastructure. Fleet refers to different types of vehicles, as shown in Figure 11. Each fleet type can be defined as individual new classes or as sub-classes of super-class *vehicle*. MaaS ontology may borrow some classes and properties from vehicle ontology, especially from the Autonomous and Connected Vehicles (ACV). Also, micromobility services, such as electric bikes and electric scooters, havegained attention in recent years and will play an important role in future smart cities' mobility. Autonomous shuttles and on-demand transit are the other types of MaaS services, usually addressing the problems with which the current urban transit agencies are faced, such as the first/last mile problem.

As shown in Figure 11, MaaS infrastructure includes:

- Physical infrastructures, such as parking areas devoted to MaaS vehicles in smart cities, or charging stations, for electric cars, bikes and scooters.
- Cyber infrastructures, such as mobility apps or Websites to reserve and access to the MaaS services.

While some of MaaS physical infrastructure elements, such as parking areas, have been covered in the current mobility ontologies, researchers have not yet defined ontologies for other important elements, for instance, charging stations, on-street parking spots allocated to electric vehicles, mobility apps, or cloud platform services.

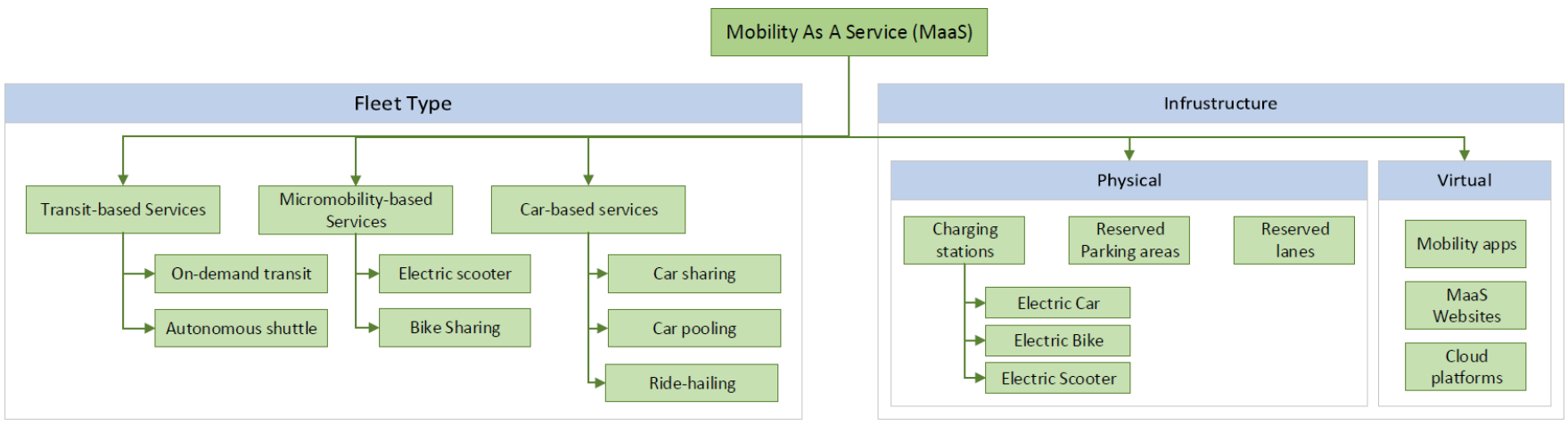

**Fig. 11** Different concepts in Mobility as a Service (MaaS).

## 8.3 Autonomous Robotics

Autonomous robotics is another domain that has rapidly advanced in recent years with potential impacts on the future of urban mobility. In mobility, robots can transport humans and goods on land, air, or water. Regarding land-based mobility, Connected and Autonomous Vehicles (CAV), such as Waymo or public transit shuttles, will be introduced to the cities in the near future. The autonomous trucks also will soon play an important role in urban and interurban logistics. As mentioned in Section 7.2.5, one of the transportation disciplines that will benefit majorly from robotics is freight transportation, where the UAVs will start delivering goods in the near future. While there are numerous advances in the robotics domain, the transportation ontologies have not included the robotics systems and autonomous vehicles. For example, Katsumi and Fox (2017b) have reviewed around 15 mobility ontologies, and none of them defined classes or properties regarding robots or autonomous vehicles.

One of the most important issues in mobility robotics, such as UAVs, is the interaction and information exchange between robots and humans or between multiple robots (Olszewska et al., 2017). In other science fields, such as Robotics and Automation, ontologies have been developed to provide knowledge representation and reasoning for autonomous robots. For example, Autonomous Robot Architecture Ontology (ROA) has been developed to explain the important concepts and their relations regarding robots architecture (Olszewska et al., 2017). The ontology defines formal concepts in the robotics domain, such as Task, Function, and Behavior, along with the spatio-temporal relationships between different objects. In the mobility domain, the interaction and information exchange between CAVs and humans, i.e., pedestrians and cyclists, or between multiple CAVs can benefit from ontology development, where ontology is used to develop intelligent decision-making systems to improve the driving safety of CAVs (Zhao et al., 2015).

Moreover, the CAVs require a sophisticated map to perceive the driving environment and make decisions. Hence, the road, pedestrian, and cycling network should be defined in more detail where the connection between them, the right of way, turns, and directions are accurately described in the ontology. Furthermore, all the sensors and cameras of CAVs should be defined in the ontology. For example, Toyota Technological Institute (TTI) team has developed three ontologies for the ADAS system: map ontology, control ontology, and vehicle ontology. These ontologies, along with a reasoning language, referred to as Semantic Web Rule Language (SWRL), have been used to analyze scenarios in the traffic ecosystem and to make driving decisions.

Furthermore, autonomous driving requires integrating different and high volume sources of data from CAVs sensors, roadside sensors, and road infrastructure. Hence, the ontology of autonomous robotics or CAVs is associated with sensors' ontology, connected roadways technologies (described in the next section), and geospatial ontology. While some efforts have been made to develop CAVs related ontologies, however, the literature lacks an ontology that incorporates all aspects of CAVs, from safety and decision-making issues to sensors and traffic data integration as well as connected roadway technologies (as explained in the next section). Hence, future smart mobility ontologies need to encompass all the aspects of CAVs, as mentioned above.

## 8.4 Connected Roadways and Internet of Vehicles Technologies

This section introduces the new technologies related to Connected Roadways and Internet of Vehicles (IoV) or Vehicular Communication Networks (VCNs)Khan et al. (2018). It helps the ontology developers to have a general view of the domain, identify the main concepts and their relationship, and how these technologies will affect the future transportation networks in smart cities.

Connected roadways refer to a set of communication devices and technologies allowing connected vehicles, travelers, and traffic management systems to communicate with each other and exchange information securely in real-time. Connected roadways mainly rely on VANETs (Vehicular ad-hoc networks). VANETs is used for four main vehicular communication (Gasmi and Aliouat, 2019), i.e.:

– Vehicle-to-Vehicle (V2V)
– Vehicle-to-Infrastructure (V2I)
– Vehicle to Personal devices (V2P) or Vehicle to Human (V2H)
– Vehicle-to-Sensor (V2S)

Vehicle-to-everything (V2X) communications are the most generic form, which covers all types of communication to everything surrounding a vehicle.

As the number of connected vehicles is increasing and will rise sharply in the future, network services to connect the huge number of vehicles need new requirements, such as robust and scalable information exchange between vehicles and other devices. Also, VANETs share many similarities with Internet Of Things (IoT) domain, which has caused researchers to give a new naming for vehicular communication technologies as Internet-of-Vehicles (IoV).

The IoV technologies can be divided into three general classes based on their coverage area range(Jawhar et al., 2018; Gasmi and Aliouat, 2019):

- Personal Area Network (PAN)
- Local Area Network (LAN)
- Wide Area Network (WAN)

As shown in Figure 12, each of these three network categories has different types of services that are briefly explained below.

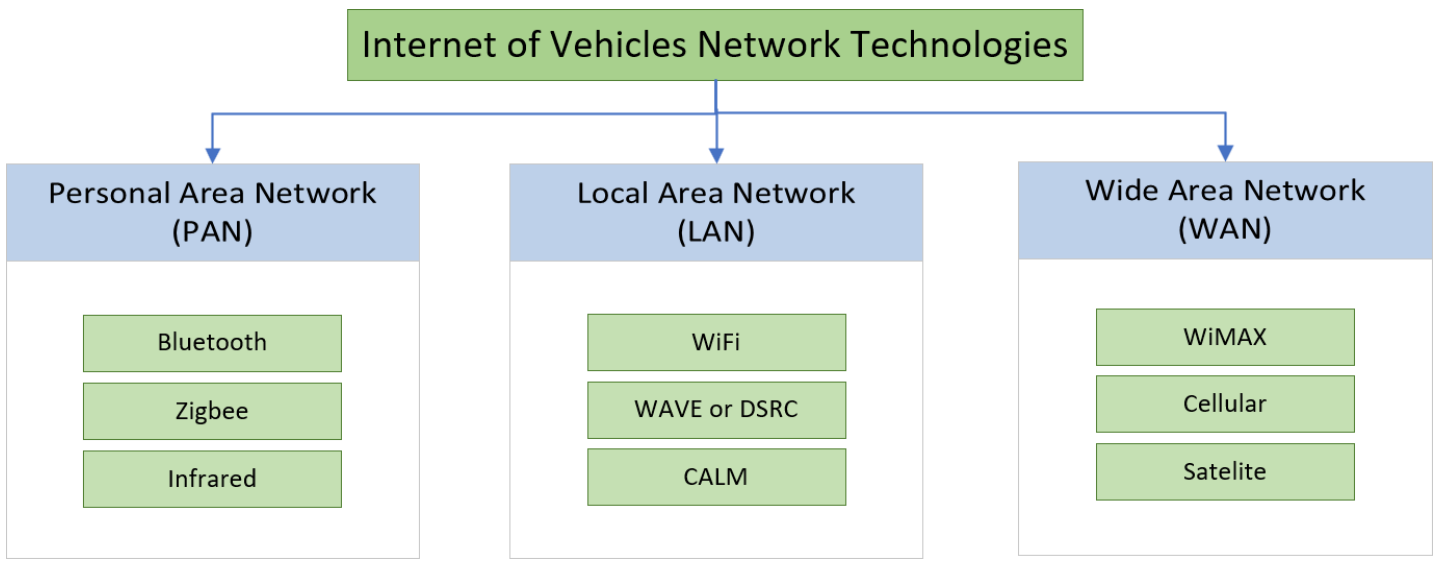

**Fig. 12** Different Network Technologies in Internet of Vehicles (IoV).

The PAN services support low bandwidth and energy consumption communications. ZigBee is a low-cost communication technology that supports short-range information exchange between a vehicle and its internal sensors (V2S). Bluetooth is also a short-range communication network that mainly supports V2P or V2H communications in many of today's vehicles.

Infrared is considered a PAN technique that transfers data at a lower rate compared to Bluetooth. However, it has some advantages, such as its large bandwidth that enables high network traffic in V2V applications. However, as infrared signals are highly affected by obstacles, their usage is limited to very short distances (Anwer and Guy, 2014).

Regarding LAN services, the most well-known technology is Wireless Access in Vehicular environments (WAVE). A WAVE system, also known as Dedicated Short-Range Communications (DSRC), refers to a system designed for efficient and reliable radio communications for V2V, V2R, or V2I direct connections (Jiang and Zhu, 2019). Generally speaking, WAVE technology is achieved via wireless access through two types of WAVE devices: vehicle on-board units (OBU) and roadside units (RSU)(Chhabra et al., 2015). RSUs are usually installed on roadside infrastructure, such as poles, road signs, or traffic signals, or electronic cabinets on the roadside. WAVE devices support a communication range of approximately 300 meters to 1000 meters for moving vehicles with speeds of 10 to 200 km/h (Chhabra et al., 2015).

WiFi technology for vehicular communication consists of roadside units, as wireless access points, to support vehicular communications inside their coverage area. WiFi services provide V2I and ad-hoc V2V communication (Zekri and Jia, 2018). WiFi technology coverage range is up to 100 m. However, it does not support vehicles moving at high speed.

Another type of LAN service is WiMAX (Worldwide Interoperability for Microwave Access). WiMAX supports vehicle communication to the Internet at a maximum distance of 50 km. It is considered as a fast and high bandwidth connection providing V2X communication.

Cellular networks can provide different vehicular communications based on radio waves at long distances (Anwer and Guy, 2014) and high mobility speeds. It includes different cellular services such as 2G, 3G, and 4G/LTE technologies that differ in their bandwidth, latency, and data transfer rate. Today's cellular networks, specifically 4G/LTE networks, are considered as a practical communication technology for V2V communications (Gasmi and Aliouat, 2019).

Satellite Communications are mainly transferring data via Microwaves or low-power radio at long distances. They can be used to set up long-range wireless networks to connect multiple cities. Moreover, satellite communication plays a crucial role in vehicle positioning in VANETs, through Global Positioning System (GPS).

Today, IoV is mainly considered as heterogeneous vehicular communications that integrate cellular, WiFi, and DSRC technologies to provide seamless and efficient communication. Moreover, with the advent of the 5G cellular networks, the performance and capability of IoV and connected roadways keep improving drastically, which consequently has given rise to new concepts and technologies in smart mobility, such as Cloud Computing, Fog Computing, and Vehicular Fog Networks (VFN).

Besides different communication technologies, IoV and connected roadways also rely on physical devices, such as sensing or actuating devices, CAVs, or RSUs, OBUs. Hence, to develop an ontology for IoV and connected roadways, one should consider other innovative ontologies, such as CAVs or mobility senors ontology. For example, the IoV technologies mentioned above are implemented through some physical RSUs or OBUs and connected to different sensors on vehicles, robots, infrastructure, or personal devices (smartphones, smartwatches, etc.). As a consequence, it is recommended that any effort to develop ontologies for innovative mobility technologies should encompass all of them, as they are mainly interrelated together.

IoV ontology development can also benefit from ontologies developed for the Internet of Things, as indeed, the concepts and technologies are pretty similar between the two domains. IoT-Lite (Bermudez-Edo et al., 2016) is an ontology developed for the IoT domain, consisting of three main classes, i.e. *objects, system, and services*. *Objects* are any entity in IoT environment. A *System* is a unit of abstraction for all the physical entities for sensing. The system has components and sub-systems. *Service* refers to any service provided by IoT devices.

The IoT-Lite ontology can be used for IoV ontology. For example, all the vehicles, RSUs, and OBUs can be defined as instances of class *Object*. Some of the objects, such as RSUs and OBUs, provide a service, for example, a DSRC service. Such devices have a *Coverage* property. The coverage of a device is of geospatial data type, which shows the area covered by a device.

Moreover, any mobility sensing device can have a *connects* property to a *service*, such as a device that provides a DSRC communication. Moreover, the ontology of transportation road networks should be connected to the IoV ontology. Also, vehicle ontology should be extended to include the innovative OBUs and other sensors used for V2V, V2I, V2P, or V2S communications.

While ontology studies have used VANET data in some specific mobility applications, such as vehicle routing (Chhabra et al., 2015), the concepts and entities in connected roadways and IoT technologies have not yet included in transportation ontology in the context of smart cities. Future efforts need to focus on defining concepts and entities in IoV and sensors technologies and defining them according to future connected roadways requirements.

## 9 Conclusion

This chapter aims to give a general overview of ontologies in the domain of smart mobility, which will be of interest to researchers and practitioners. It will be of help to not only define the current concepts and entities in today's cities, but also to include the innovative and disruptive mobility technologies and services in the future smart cities. We explain what an ontology is and why it has gained attraction in recent years across different fields of science. Also, we cover the main components of ontology and different ontology design approaches and languages.

While there are several languages and approaches in the literature for ontology development, OWL has proven itself as the formal language for ontology development. Today, numerous mobility ontologies have been developed using OWL language, and due to its flexibility and coherence, OWL seems to be the best option to develop formal general ontologies in the domain of smart mobility. However, the transportation domain is changing fast and will keep on changing in the upcoming years. Future transportation road networks do not consist only nodes, links, and street furniture, but also include a wide spectrum of communication technologies, sensors, and autonomous vehicles. Future streets and highways will not differ only in the number of lanes or speed limit, but also vary depending on different V2X communication technologies they offer, various types of on-road or road-side sensors they have. Hence, future road network ontologies need to be extended considerably to cover these innovative technologies. Moreover, the current mobility ontologies, such as freight transportation, vehicle ontology, or transit system ontology, will change considerably due to the different types of autonomous passenger and freight vehicles that will surge to the urban environments in the near future. Besides the effects on physical and cyber networks, many concepts in the current transportation domain, such as vehicle ownership, driving behavior, parking requirements, etc., will vary dramatically due to the disruptive technologies, which consequently call for new ontology development efforts.

Considering the fact that different innovative technologies finally interact with each other in future smart cities environments, stress the importance of a ”general” smart mobility ontology that reflects the effects of future technological disruptions in the domain of transport. Currently, some applications or core ontologies have been developed according to the different aspects of connected and autonomous vehicles, for example, for safe autonomous driving. However, no general mobility ontology, in the context of smart cities, have yet included such innovative technologies. Moreover, as mentioned by Katsumi and Fox (2017b), only when an ontology is well documented, maintained, and accessible on the Web is it reusable by other domain experts. Hence, any ontology development effort not only should deploy extensive and relevant domain expertise, but also requires choosing an appropriate approach, following standard procedures, and is written in a standard language using the right ontology editor. Hence, the chapter can serve as a useful starting point for transportation researchers and ontology developers who aim to generate a formal general mobility ontology in the context of smart cities.

## References

(2014) INSPIRE Data Specification on Transport Networks – Technical Guidelines. https://inspire.ec.europa.eu/file/1723/download?token=0GOYYbMF

(2017) GTFS Realtime Overview. https://developers.google.com/transit/gtfs-realtime/

Al-Sayed MM, Hassan HA, Omara FA (2019) Towards evaluation of cloud ontologies. Journal of Parallel and Distributed Computing 126:82–106

Anand N, van Duin R, Tavasszy L (2014) Ontology-based multi-agent system for urban freight transportation. International Journal of Urban Sciences 18(2):133–153

Anwer MS, Guy C (2014) A survey of vanet technologies. Journal of Emerging Trends in Computing and Information Sciences 5(9):661–671

Arp R, Smith B, Spear AD (2015) Building Ontologies with Basic Formal Ontology. The MIT Press

Badu-Marfo G, Farooq B, Patterson Z (2019) A perspective on the challenges and opportunities for privacy-aware big transportation data. Journal of Big Data Analytics in Transportation 1(1):1–23

Bellini P, Benigni M, Billero R, Nesi P, Rauch N (2014) Km4city ontology building vs data harvesting and cleaning for smart-city services. Journal of Visual Languages & Computing 25:827–839

Bermudez-Edo M, Elsaleh T, Barnaghi P, Taylor K (2016) Iot-lite: a lightweight semantic model for the internet of things. In: 2016 Intl IEEE Conferences on Ubiquitous Intelligence & Computing, Advanced and Trusted Computing, Scalable Computing and Communications, Cloud and Big Data Computing, Internet of People, and Smart World Congress (UIC/ATC/ScalCom/CBDCom/IoP/SmartWorld), IEEE, pp 90–97

Berners-Lee T (2007) ConverterToRdf

Chen W, Kloul L (2018) An ontology-based approach to generate the advanced driver assistance use cases of highway traffic. In: KEOD, pp 73–81

Chhabra S, Bali RS, Kumar N (2015) Dynamic vehicle ontology based routing for vanets. Procedia Computer Science 57:789–797
Choi C, Cho M, Kang Ey, Kim P (2006) Travel ontology for recommendation system based on semantic web. In: 2006 8th International Conference Advanced Communication Technology, IEEE, vol 1, pp 624–627
Codescu M, Horsinka G, Kutz O, Mossakowski T, Rau R (2011) Osmonto-an ontology of openstreetmap tags. State of the map Europe (SOTM-EU) 2011
Corsar D, Markovic M, Gault PE, Mehdi M, Edwards P, Nelson JD, Cottrill CD, Sripada S (2015) Travelbot: journey disruption alerts utilising social media and linked data. In: Proceedings of the Posters and Demonstrations Track of the 14th International Semantic Web Conference (ISWC 2015), CEUR-WS
Desmond Mogotlane K, Fonou-Dombeu JV (2016) Automatic conversion of relational databases into ontologies: A comparative analysis of protege plug-ins performances. International journal of Web & Semantic Technology 7(3/4):21–40, DOI 10.5121/ijwest.2016.7403
Falquet G, Métral C, Teller J, Tweed C (2011) Ontologies in urban development projects. Springer Science & Business Media
Fensel D, Van Harmelen F, Horrocks I, McGuinness DL, Patel-Schneider PF (2001) Oil: An ontology infrastructure for the semantic web. IEEE intelligent systems 16(2):38–45
Gasmi R, Aliouat M (2019) Vehicular ad hoc networks versus internet of vehicles-a comparative view. In: 2019 International Conference on Networking and Advanced Systems (ICNAS), IEEE, pp 1–6
Genesereth M, Fikes R (1992) Knowledge interchange format-version 3.0: Reference manual. Stanford University
Gruber TR, et al. (1993) A translation approach to portable ontology specifications. Knowledge acquisition 5(2):199–220
Gutierrez C, Hurtado CA, Vaisman A (2006) Introducing time into rdf. IEEE Transactions on Knowledge and Data Engineering 19(2):207–218
Hobbs JR, Pan F (2006) Time ontology in owl. W3C working draft 27:133
Hogan A (2020a) Resource Description Framework, Springer International Publishing, Cham, pp 59–109. DOI 10.1007/978-3-030-51580-5 3
Hogan A (2020b) Web ontology language. In: The Web of Data, Springer, pp 185–322
Janowicz K, Haller A, Cox SJ, Le Phuoc D, Lefrançois M (2019a) Sosa: A lightweight ontology for sensors, observations, samples, and actuators. Journal of Web Semantics 56:1–10
Janowicz K, Haller A, Cox SJ, Le Phuoc D, Lefrançois M (2019b) Sosa: A lightweight ontology for sensors, observations, samples, and actuators. Journal of Web Semantics 56:1 – 10, DOI https://doi.org/10.1016/j.websem.2018.06.003
Jawhar I, Mohamed N, Al-Jaroodi J (2018) Networking architectures and protocols for smart city systems. Journal of Internet Services and Applications 9(1):26

Jiang R, Zhu Y (2019) Wireless Access in Vehicular Environment, Springer International Publishing, Cham, pp 1–5

Kalibatiene D, Vasilecas O (2011) Survey on ontology languages. In: International Conference on Business Informatics Research, Springer, pp 124–141

Katsumi M, Fox M (2017a) icity ontology initial release

Katsumi M, Fox M (2017b) icity ontology version 1.0 report. University of Toronto Transportation Research Institute

Khan AA, Abolhasan M, Ni W (2018) 5g next generation vanets using sdn and fog computing framework. In: 2018 15th IEEE Annual Consumer Communications & Networking Conference (CCNC), IEEE, pp 1–6

Klein LA (2017) ITS Sensors and Architectures for Traffic Management and Connected Vehicles. CRC Press

Klotz B, Troncy R, Wilms D, Bonnet C (2018) Vsso: The vehicle signal and attribute ontology. In: SSN@ ISWC, pp 56–63

Krieger HU (2008) Where temporal description logics fail: Representing temporally-changing relationships. In: Annual Conference on Artificial Intelligence, Springer, pp 249–257

Krieger HU (2010) A general methodology for equipping ontologies with time. In: LREC

Kumar VRS, Khamis A, Fiorini S, Carbonera JL, Alarcos AO, Habib M, Goncalves P, Li H, Olszewska JI (2019) Ontologies for industry 4.0. The Knowledge Engineering Review 34

Lorenz B, Ohlbach HJ, Yang L (2005) Ontology of transportation networks. Citeseer

Mcguinness DL, Fikes R, Hendler J, Stein LA (2002) Daml+oil: an ontology language for the semantic web. IEEE Intelligent Systems 17(5):72–80, DOI 10.1109/MIS.2002.1039835

Mejhed Mkhinini M, Labbani-Narsis O, Nicolle C (2020) Combining uml and ontology: An exploratory survey. Computer Science Review 35:100,223, DOI https://doi.org/10.1016/j.cosrev.2019.100223

Métral C, Cutting-Decelle AF (2011) Ontologies fundamentals. In: Ontologies in Urban Development Projects, Springer, pp 105–122

Noura M, Gyrard A, Heil S, Gaedke M (2019) Automatic knowledge extraction to build semantic web of things applications. IEEE Internet of Things Journal 6(5):8447–8454

Noy NF, Fergerson RW, Musen MA (2000) The knowledge model of protege-2000: Combining interoperability and flexibility. In: International Conference on Knowledge Engineering and Knowledge Management, Springer, pp 17–32

Noy NF, McGuinness DL, et al. (2001) Ontology development 101: A guide to creating your first ontology

Obrst L, Grüninger M, Baclawski K, Bennett M, Brickley D, Berg-Cross G, Hitzler P, Janowicz K, Kapp C, Kutz O, et al. (2014) Semantic web and big data meets applied ontology. Applied Ontology

Olszewska JI, Barreto M, Bermejo-Alonso J, Carbonera J, Chibani A, Fiorini S, Goncalves P, Habib M, Khamis A, Olivares A, et al. (2017) Ontology for autonomous robotics. In: 2017 26th IEEE International Symposium on Robot and Human Interactive Communication (RO-MAN), IEEE, pp 189–194

Roussey C, Pinet F, Kang MA, Corcho O (2011) An introduction to ontologies and ontology engineering. In: Ontologies in Urban development projects, Springer, pp 9–38

Smith B (1998) An introduction to ontology. National Center for Geographic Information and Analysis

Verstichel S, Ongenae F, Loeve L, Vermeulen F, Dings P, Dhoedt B, Dhaene T, De Turck F (2011) Efficient data integration in the railway domain through an ontology-based methodology. Transportation Research Part C: Emerging Technologies 19(4):617–643

Viktorović M, Yang D, Vries Bd (2020) Connected traffic data ontology (ctdo) for intelligent urban traffic systems focused on connected (semi) autonomous vehicles. Sensors 20(10):2961

Welty C, Guarino N (2001) Supporting ontological analysis of taxonomic relationships. Data & Knowledge Engineering 39(1):51 – 74, DOI https://doi.org/10.1016/S0169-023X(01)00030-1

Welty C, Fikes R, Makarios S (2006) A reusable ontology for fluents in owl. In: FOIS, vol 150, pp 226–236

Zekri A, Jia W (2018) Heterogeneous vehicular communications: A comprehensive study. Ad Hoc Networks 75:52–79

Zhao L, Ichise R, Mita S, Sasaki Y (2015) Core ontologies for safe autonomous driving. In: International Semantic Web Conference (Posters & Demos)